\documentclass[a4paper]{article}

\usepackage[english]{babel}
\usepackage[utf8]{inputenc}
\usepackage[colorinlistoftodos]{todonotes}
\usepackage{amsmath}
\usepackage{amssymb}
\usepackage{amsthm} 
\usepackage{mathtools}
\usepackage{mathrsfs}
\usepackage{siunitx}
\usepackage{subcaption}

\usepackage{hyperref}
\usepackage{algpseudocode}

\usepackage{tikz}

\usepackage{tabularx}
\usepackage{booktabs}
\usepackage{multirow}

\usepackage{empheq}

\usepackage{lipsum}

\usepackage[toc,page]{appendix}

\usepackage{rotating}

\usepackage{makecell}

\usepackage{lscape}
\usepackage{xcolor,colortbl}
\definecolor{lavender}{rgb}{0.9, 0.9, 0.98}

\theoremstyle{plain}
\theoremstyle{plain}
\theoremstyle{plain}
\theoremstyle{plain}
\theoremstyle{plain}
\theoremstyle{definition}
\theoremstyle{remark}

\everymath{\displaystyle}
\usepackage[a4paper, total={6in, 8in}]{geometry}
\usepackage{csquotes}
\usepackage[backend=bibtex,style=numeric-comp,giveninits=true,doi=false,isbn=false,url=false,eprint=false,maxbibnames=99]{biblatex}

\newif\ifdouble

\newcommand{\papertitle}{Liquid Fourier Latent Dynamics Networks for  fast \\ GPU-based numerical simulations in computational cardiology}

\newcommand{\keywordOne}{Liquid Fourier Latent Dynamics Networks}
\newcommand{\keywordTwo}{Scientific Machine Learning}
\newcommand{\keywordThree}{Computational cardiology}
\newcommand{\keywordFour}{Digital twins}

\newcommand{\ANNRhs}{\mathcal{B L \kern0.05em N \kern-0.05em M}}
\newcommand{\NN}{\mathcal{N \kern-0.05em N}}

\newcommand{\NumLNNState} {N_y}

\newcommand{\NumANNWeights} {N_\mathrm{w}}

\newcommand{\IdentityVec}{\mathbb{I}}

\newcommand{\fZero}{{\boldsymbol{f}_0}}

\newcommand{\Iion}{{\mathcal{I}_{\mathrm{ion}}}}
\newcommand{\Iapp}{{\mathcal{I}_{\mathrm{app}}}}
\newcommand{\DiffTens}{\boldsymbol{D}_{\mathrm{M}}}
\newcommand{\Pot}{u}

\newcommand{\Ionic}{\boldsymbol{y}}

\newcommand{\RhsIonic}{\boldsymbol{F}}

\newcommand{\GCaL}{G_\mathrm{CaL}}
\newcommand{\GNa}{G_\mathrm{Na}}
\newcommand{\GKr}{G_\mathrm{Kr}}

\newcommand{\Da}{D_\mathrm{ani}}
\newcommand{\Di}{D_\mathrm{iso}}
\newcommand{\Dp}{D_\mathrm{purk}}
\newcommand{\tLVstim}{t_\mathrm{LV}^\mathrm{stim}}

\newcommand{\paramsgeneric}{\boldsymbol{w}}

\newcommand{\spaceField}{\boldsymbol{G}}
\newcommand{\spaceTimeField}{\boldsymbol{D}}
\newcommand{\fourier}{\boldsymbol{B}}
\newcommand{\stateLNN}{\boldsymbol{y}}
\newcommand{\stateROM}{\boldsymbol{s}}

\newcommand{\inputSign}{\boldsymbol{I}}

\newcommand{\outputField}{\boldsymbol{u}}

\newcommand{\spacevar}{\boldsymbol{x}}
\newcommand{\timevar}{t}

\newcommand{\spaceDomain}{\Omega}
\newcommand{\timeMax}{T}

\newcommand{\SETinputSign}{\mathcal{I}}

\newcommand{\SEToutputField}{\mathcal{U}}

\newcommand{\NUMstateROM}{N_s}

\newcommand{\NUMinputSign}{N_I}

\newcommand{\NNgeneric}{\mathcal{N\!N}}

\newcommand{\ROMrhsliquid}{\NNgeneric_{\mathrm{dyn}}^{\mathrm{CfC, NCP}}}
\newcommand{\ROMrhsNCP}{\NNgeneric_{\mathrm{dyn}}^{\mathrm{NCP}}}
\newcommand{\ROMrhs}{\NNgeneric_{\mathrm{dyn}}}
\newcommand{\ROMobs}{\NNgeneric_{\mathrm{rec}}}

\newcommand{\paramsROMf}{\paramsgeneric_f^i}
\newcommand{\paramsROMg}{\paramsgeneric_g^i}
\newcommand{\paramsROMh}{\paramsgeneric_h^i}
\newcommand{\paramsROMdyn}{\paramsgeneric_{\mathrm{dyn}}}
\newcommand{\paramsROMobs}{\paramsgeneric_{\mathrm{rec}}}

\graphicspath{{pictures/}}
\doublefalse

\title{{\papertitle}}
\author{Matteo Salvador$^{1, 2, 3, *}$, Alison Lesley Marsden$^{1, 2, 3, 4}$}

\date{\footnotesize
    $^1$ Institute for Computational and Mathematical Engineering, Stanford University, CA, USA \\
    $^2$ Cardiovascular Institute, Stanford University, CA, USA \\
    $^3$ Pediatric Cardiology, Stanford University, CA, USA \\
    $^4$ Department of Bioengineering, Stanford University, CA, USA \\
    $^*$ \textit{Corresponding author} (\texttt{msalvad@stanford.edu}) \\
}

\begin{document}
	\maketitle
	
	\begin{abstract}
		Scientific Machine Learning (ML) is gaining momentum as a cost-effective alternative to physics-based numerical solvers in many engineering applications. In fact, scientific ML is currently being used to build accurate and efficient surrogate models starting from high-fidelity numerical simulations, effectively encoding the parameterized temporal dynamics underlying Ordinary Differential Equations (ODEs), or even the spatio-temporal behavior underlying Partial Differential Equations (PDEs), in appropriately designed neural networks.
We propose an extension of Latent Dynamics Networks (LDNets), namely Liquid Fourier LDNets (LFLDNets), to create parameterized space-time surrogate models for multiscale and multiphysics sets of highly nonlinear differential equations on complex geometries.
LFLDNets employ a neurologically-inspired, sparse, liquid neural network for temporal dynamics, relaxing the requirement of a numerical solver for time advancement and leading to superior performance in terms of tunable parameters, accuracy, efficiency and learned trajectories with respect to neural ODEs based on feedforward fully-connected neural networks.
Furthermore, in our implementation of LFLDNets, we use a Fourier embedding with a tunable kernel in the reconstruction network to learn high-frequency functions better and faster than using space coordinates directly as input.
We challenge LFLDNets in the framework of computational cardiology and evaluate their capabilities on two 3-dimensional test cases arising from multiscale cardiac electrophysiology and cardiovascular hemodynamics.
This paper illustrates the capability to run Artificial Intelligence-based numerical simulations on single or multiple GPUs in a matter of minutes and represents a significant step forward in the development of physics-informed digital twins.
	\end{abstract}
	
        \newpage
	\section{Introduction}
\label{sec:introduction}

Physics-informed Machine Learning (ML) \cite{Karniadakis2021}, and indeed scientific ML more broadly, is revolutionizing many disciplines, ranging from mechanical and aerospace engineering \cite{Ferrari2024, Willcox2024} to computational medicine \cite{Laubenbacher2024} by strategically combining physical governing principles, scientific computing techniques, and fast data-driven approaches.
Recent advances in operator learning \cite{Azizzadenesheli2024}, such as Deep Operator Networks \cite{Goswami2023, He2024, Howard2023, Lu2021} Fourier and Graph Neural Operators \cite{Li2020, Li2021}, General Neural Operator Transformers \cite{Hao2023}, Branched Latent Neural Maps \cite{Salvador2024BLNM} and Latent Dynamics Networks (LDNets) \cite{Regazzoni2024}, enable the creation of fast and accurate surrogate models starting from a set of numerical simulations generated via high-performance computing after performing numerical discretization of Ordinary/Partial Differential Equations (ODEs/PDEs).

In principle, these surrogate models reproduce the temporal/spatio-temporal behavior of dynamical systems described by differential equations, potentially spanning multiple PDE systems \cite{Rahman2024} and geometric variations \cite{Li2023} at the cost of a reasonably small approximation error.
Nevertheless, they usually encode a limited set and variability of model parameters, initial and boundary conditions, presented in test cases computed in relatively simple computational domains in fairly regular meshes.

LDNets, introduced in Regazzoni et al. \cite{Regazzoni2024}, define a novel architecture to create surrogate models for parameterized physical processes, such as those arising from systems of differential equations.
LDNets jointly discover a low-dimensional nonlinear manifold while learning the spatio-temporal system dynamics, avoiding operations in the high-dimensional space.
LDNets achieve higher accuracy on highly nonlinear mathematical problems with significantly fewer trainable parameters compared to both intrusive and non-intrusive state-of-the-art methods based on proper orthogonal decomposition, autoencoders, and various types of recurrent neural networks \cite{Regazzoni2024}.

In this paper, we propose an extension of LDNets, namely Liquid Fourier LDNets (LFLDNets), as an effective operator learning method that demonstrates superior performance in terms of number of tunable parameters, expressiveness, interpretability, overall accuracy and computational efficiency during training and inference.
LFLDNets use a simple and compact Liquid Neural Network (LNN) instead of a feedforward Fully-Connected Neural Network (FCNN) to track the temporal dynamics of a set of states, capturing the global behavior of highly nonlinear parameterized multiscale PDEs.
Furthermore, they use a Fourier embedding with a learnable kernel matrix for space coordinates, so that a second FCNN reconstructs the entire space-time dynamics of the PDE system by properly capturing complex functions.

We demonstrate the capabilities of LFLDNets on two challenging 3D patient-specific cases arising in computational cardiology \cite{Niederer2019Nature}.
We first build multiscale surrogate models for cardiac electrophysiology \cite{Salvador2024DTCHD} in a complex pediatric congenital heart disease model and second simulate cardiovascular hemodynamics in a healthy pediatric aorta \cite{Updegrove2017}.
These geometries are spatially discretized with a fine unstructured tetrahedral grid; the hemodynamics case also contains anisotropic element refinement within the boundary layer.
Furthermore, we demonstrate that even in these complex scenarios, LFLDNets allow for time steps that are orders of magnitude larger than those commonly used in physics-based simulations due to the constraints introduced by the specific numerical discretization scheme.

This work introduces a new scientific ML tool, namely LFLDNets, and demonstrates its potential in two realistic examples from computational medicine.
During inference, LFLDNets produce, within some tolerance, results equivalent to 3D space-time cardiovascular simulations in a matter of minutes on one or multiple Graphics Processing Units (GPUs) while spanning different sets of model parameters, initial, and boundary conditions.
This tool defines another contribution to building complex multiscale and multiphysics spatio-temporal surrogate models to create accurate, efficient and certified digital replicas for cardiovascular applications \cite{CorralAcero2020, Fedele2023, Gerach2021, Jung2022, Gillette2021, Salvador2024DTCHD, Salvador2024EMROM4CH, Viola2023}.
	\section{Methods}
\label{sec:methods}

Here, we introduce LFLDNets and highlight the key differences between these and LDNets.
We explain how LNNs work and discuss their advantages compared to neural ODEs based on FCNNs.
We also show the importance of adding a proper Fourier embedding for the space coordinates to better learn high-frequency functions.

Then, we present the mathematical governing equations and numerical formulations for two challenging test cases in cardiac electrophysiology and cardiovascular hemodynamics.

\subsection{Liquid Fourier Latent Dynamics Networks}
\label{sec:methods:LFLDNets}

\begin{figure}[t!]
    \includegraphics[width=1.0\textwidth]{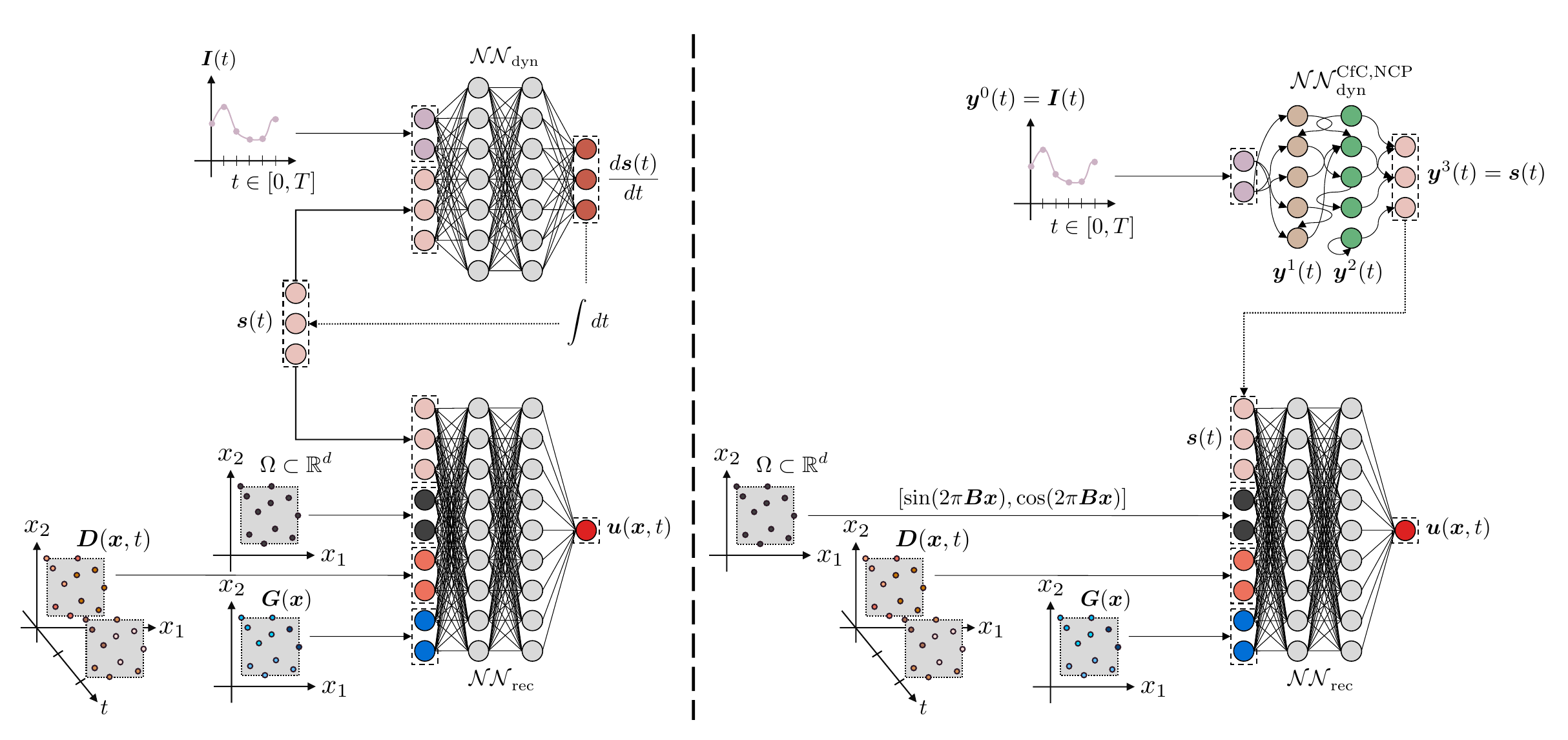}
    \caption{Comparison between Latent Dynamics Networks (left) and Liquid Fourier Latent Dynamics Networks (right).}
   \label{fig:LDNets_LFLDNets}
\end{figure}

We compare the architectures of LDNets and LFLDNets in Figure~\ref{fig:LDNets_LFLDNets}.
The general mathematical formulation for LFLDNets reads:

\begin{equation*}\label{eqn:ROM}
    \left\{
    \begin{aligned}
        \stateROM(\timevar) &= \ROMrhsliquid(\inputSign(\timevar); \paramsROMdyn), & & \text{for $\timevar \in (0, \timeMax]$}, \\
        \stateROM(0) &= \stateROM_0, \\
        \outputField(\spacevar, \timevar) &= \ROMobs(\stateROM(\timevar), [\cos(2 \pi \fourier \spacevar), \sin(2 \pi \fourier \spacevar)]^T, \spaceField(\spacevar), \spaceTimeField(\spacevar, t); \paramsROMobs),
        & & \text{for $\spacevar \in \spaceDomain^i$, $\timevar \in [0, \timeMax]$,}
    \end{aligned}
    \right.
\end{equation*}
for $i = \{ 1, ..., N_{\spaceDomain} \}$, where $N_{\spaceDomain}$ defines the total number of computational domains.
As with LDNets \cite{Regazzoni2024}, LFLDNets are made of two neural networks.
The first, called the dynamics network $\ROMrhsliquid$ (with tunable parameters $\paramsROMdyn$), accounts for the temporal evolution of a vector $\stateROM(\timevar) \in \mathbb{R}^{\NUMstateROM}$ that tracks the global state of a set of PDEs, starting from an initial condition $\stateROM(0)=\stateROM_0$, in the time interval $(0, T]$.
The second, called the reconstruction network $\ROMobs$ (with trainable weights and biases $\paramsROMobs$), operates in the space domain and reconstructs space-time variables, denoted as $\outputField(\spacevar, \timevar)$, which can be any output extracted from a system of PDEs.

LFLDNets are neural operators that map time-dependent $\inputSign(t) \in \SETinputSign$, space-dependent $\spaceField(\spacevar) \in \mathcal{G}$ and space-time $\spaceTimeField(\spacevar, t) \in \mathcal{D}$ input signals coming from physics-based model parameters, coefficients, forcing terms, initial and boundary conditions extracted from a set of differential equations, as well as geometrical parameters associated with one or multiple computational domains $\spaceDomain^i \subset \mathbb{R}^{N_d}$, such as those arising from statistical shape modeling \cite{Rodero2021} or signed distance fields \cite{Kong2023,Verhulsdonk2024}, into a space-time output field $\outputField(\spacevar, \timevar) \in \SEToutputField$.
The matrix $\fourier \in \mathbb{R}^{N_f \times N_d}$ represents a trainable Fourier encoding for the space coordinates $\spacevar \in \spaceDomain^i \subset \mathbb{R}^{N_d}$, where $N_f$ is the total number of frequencies.
We remark that scalar and vector model parameters can be reinterpreted as input signals that are constant in space and time \cite{Regazzoni2024, Salvador2024BLNM}.

Compared to LDNets, LFLDNets present two key novelties, residing in the use of LNNs and Fourier embedding, described below.

\subsubsection{Liquid Neural Networks}
\label{sec:methods:LNN}

LNNs are lightweight and flexible architectures, that remain adaptable after the training phase, with excellent generalization properties, even in the presence of scarce data and a significant amount of noise, by using a limited number of expressive biologically-inspired neurons and corresponding connections \cite{Hasani2021LNN, Lechner2020NCP, Hasani2022CfC}.
A similar trend in developing powerful, explainable and tiny ML models encoding the behavior of complex dynamical systems can be also observed in other recent works \cite{Liu2024, Regazzoni2024, Salvador2024DTCHD, Salvador2024EMROM4CH}.

In this section we review the mathematical framework of how LNNs evolved, moving from Liquid-Time Constant (LTC) networks \cite{Hasani2021LNN} to neural circuit policies (NCPs) \cite{Lechner2020NCP} and closed-form continuous-time (CfC) networks \cite{Hasani2022CfC}.
We also motivate why these architectures can lead to an improved design of LDNets \cite{Regazzoni2024}.

The first implementation of LNNs was built from neural ODEs \cite{Chen2019}, which model how dynamical systems evolve by introducing a neural network-based right hand side for a system of ODEs.
Specifically, in neural ODEs, the time derivative of a hidden state vector $\stateLNN(\timevar) \in \mathbb{R}^{\NumLNNState}$ can be expressed as follows \cite{Hasani2021LNN}:

\begin{equation}
\dfrac{d\stateLNN(\timevar)}{d\timevar} = \ROMrhs(\timevar, \stateLNN(\timevar), \inputSign(\timevar); \paramsROMdyn),
\label{eq:NODE}
\end{equation}
where $\ROMrhs \colon \mathbb{R}^{1 + \NumLNNState + \NUMinputSign} \to \mathbb{R}^{\NumLNNState}$ represents a FCNN, with weights and biases encoded in $\paramsROMdyn \in \mathbb{R}^{\NumANNWeights}$, vector $\inputSign(\timevar) \in \SETinputSign$ defines some time-dependent, exogenous, presynaptic input signals laying in a suitable functional space.

LNNs in their first, LTC version, introduced a different expression of the right hand side comprising a liquid time-constant parameter vector $\boldsymbol{w}_{\tau} \in \mathbb{R}^{\NumLNNState}$ and an additional tunable parameter vector $\boldsymbol{A} \in \mathbb{R}^{\NumLNNState}$, that is:


\begin{equation}
\dfrac{d\stateLNN(\timevar)}{d\timevar} = -\boldsymbol{w}_{\tau} \odot \stateLNN(\timevar) + \boldsymbol{S}(t)  = -\boldsymbol{w}_{\tau} \odot \stateLNN(\timevar) + \ROMrhs(\timevar, \stateLNN(\timevar), \inputSign(\timevar); \paramsROMdyn) \odot (\boldsymbol{A} - \stateLNN(\timevar)),
\label{eq:LTC}
\end{equation}
where $\odot$ is the Hadamard product.
The term $\boldsymbol{S}(t)$ represents a conductance-based synapse model expressed using the Hodgkin-Huxley formalism \cite{Hodgkin1952}.
Solving this set of ODEs rather than Equation~\eqref{eq:NODE} offers several advantages, such as bounded dynamics for arbitrary sets of inputs, stability and superior expressivity \cite{Hasani2021LNN}.

LTC networks still use FCNNs within $\ROMrhs$.
On the other hand a NCP, which is built on top of LTC networks, replaces the FCNN in Equation~\eqref{eq:LTC} with a sparse recurrent neural network that is inspired by the nervous system of the Caenorhabditis elegans nematode \cite{Meyer2008}.
NCPs show higher generalizability, interpretability and robustness compared with orders-of-magnitude larger black-box ML models \cite{Lechner2020NCP}, and can learn causal relationships directly from data \cite{Vorbach2021LNN}.
As shown in Figure~\ref{fig:LDNets_LFLDNets} (right), the NCP dynamics network $\ROMrhsNCP$ is made up of four layers: the first input layer, where we find the sensory neurons $\stateLNN^0(\timevar) = \inputSign(\timevar)$, the second and third layers, comprised of the inter neurons $\stateLNN^1(\timevar)$ and command neurons $\stateLNN^2(\timevar)$, respectively, and the four output layer, defined by the motor neurons $\stateLNN^3(\timevar) = \stateROM(\timevar)$ \cite{Lechner2020NCP}.
For further details about NCPs and their wiring algorithm we refer to \cite{Lechner2020NCP}.

Eventually, CfC networks $\ROMrhsliquid$, built on top of LTC networks with NCP, solve the dynamical system modelled in Equation~\eqref{eq:LTC} by using a closed-form solution, which reads \cite{Hasani2022CfC}:

\begin{equation}
\begin{aligned}
\stateLNN(\timevar) = (\stateLNN_0 - \boldsymbol{A}) \odot e^{-\left[ \boldsymbol{w}_{\tau} + \ROMrhsliquid(\stateLNN(\timevar), \inputSign(\timevar); \paramsROMdyn) \right] \timevar} \odot \ROMrhsliquid(-\stateLNN(\timevar), -\inputSign(\timevar); \paramsROMdyn) + \boldsymbol{A},
\end{aligned}
\label{eq:CfC}
\end{equation}
where $\stateLNN(\timevar) = [\stateLNN^1(\timevar), \stateLNN^2(\timevar), \stateLNN^3(\timevar)]^T$, i.e. a vector containing all the neurons except the sensory ones.
Specifically, CfC networks have an explicit time dependence in their formulation that, in contrast to other methods based on neural ODEs \cite{Chen2019, Dupont2019, regazzoni2019modellearning, Regazzoni2024}, does not require a numerical solver with a specific time stepping scheme to obtain the temporal rollout.
This closed-form solution also allows one to query $\stateLNN(\timevar)$ at arbitrary times $t$, which do not need to be uniformly distributed in $[0, T]$.
However, training the dynamics network $\ROMrhsliquid$ in Equation~\eqref{eq:CfC} presents several difficulties, which are mostly driven by vanishing/exploding gradient issues related to its exponential term.
This leads to the final closed-form solution that propagates the input signals $\stateLNN^0(\timevar) = \inputSign(\timevar)$ to the output state $\stateLNN^3(\timevar) = \stateROM(\timevar)$ of the CfC, NCP dynamics network, which reads:

\begin{equation}
\begin{split}
\stateLNN^{i+1}(\timevar) = &\sigma(-f(\stateLNN^{i}(\timevar); \paramsROMf) \boldsymbol{t}) \odot g(\stateLNN^{i}(\timevar), \inputSign(\timevar); \paramsROMg) + \\
& + \left[ 1 - \sigma(-f(\stateLNN^{i}(\timevar); \paramsROMf) \boldsymbol{t}) \right] \odot h(\stateLNN^{i}(\timevar); \paramsROMh) \;\;\; \text{for} \;\;\; i = 0, ..., 2
\label{eq:CfCFinal}
\end{split}
\end{equation}
where we remark that $\ROMrhsliquid$ encodes the tunable parameters $\paramsROMdyn = [\paramsROMf, \paramsROMg, \paramsROMh]^T$ for $i = 0, ..., 2$.

The model expressed in Equation~\eqref{eq:CfCFinal} is anticipated by a mixed long short-term memory (LSTM) architecture \cite{Hochreiter1997}, which better captures long-term dependencies while avoiding gradient issues.
Overall, the CfC, NCP dynamics network is proven to be at least an order of magnitude faster than its ODE-based counterpart \cite{Hasani2022CfC}. Furthermore, CfCs are as expressive as neural ODEs, while requiring fewer tunable parameters, thus maximizing the trade-off between accuracy and efficiency \cite{Hasani2022CfC}.

We use the LeCun hyperbolic tangent activation function for all layers of the $\ROMrhsliquid$ dynamics network.
Moreover, we do not employ any backbone units, that is we learn $f$, $g$ and $h$ separately without any shared representations \cite{Hasani2022CfC}.

\subsubsection{Fourier encoding}
\label{sec:methods:Fourier}

Instead of passing the space coordinates $\spacevar \in \spaceDomain \subset \mathbb{R}^d$ as inputs to the reconstruction network directly, we consider the following Fourier embedding \cite{Hennigh2020}:

\begin{equation*}
\spacevar \mapsto [\cos(2 \pi \fourier \spacevar), \sin(2 \pi \fourier \spacevar)]^T,
\label{eq:fourier}
\end{equation*}
where $\fourier \in \mathbb{R}^{N_f \times N_d}$ is a trainable encoding, being $N_f$ the number of frequencies.
Specifically, adding Fourier features enables learning high-frequency functions faster and easier than using space coordinates, especially from complex space-time outputs exhibiting steep gradients and sharp variations \cite{Tancik2020, Sitzmann2020}.
Differently from \cite{Hennigh2020, Sitzmann2020}, where Fourier Feature Networks or Sinusoidal Representation Networks are employed, we do not consider any modifications in the structure of our feedforward fully-connected reconstruction network.
Furthermore, differently from \cite{Regazzoni2024}, we use the continuously differentiable Gaussian Error Linear Unit (GELU) activation function \cite{Hendrycks2016} in the reconstruction network.

\subsubsection{Training and inference}
\label{sec:methods:HPO}

In each test case, we perform hyperparameter tuning to automatically find an optimal configuration for LFLDNets.
The hyperparameters include the number of layers and neurons per layer of $\ROMobs$, the number of frequencies $N_f$ for the Fourier embedding, the total number of neurons for $\ROMrhsliquid$ (as the number of layers is fixed to 4), and the total number of states $\NUMstateROM$.
We use the Tree-structured Parzen Estimator (TPE) Bayesian algorithm to tune these hyperparameters \cite{Bergstra2011, Optuna2019}.
We fix the total number of trials to 20.

We simultaneously train multiple neural networks using 4 Nvidia A40 GPUs of one supercomputer node available at the Stanford Research Computing Center.
This node is equipped with 250 GB of RAM, so that the entire dataset of parameterized spatio-temporal numerical simulations can be loaded once during setup and specific portions are sent to the GPUs.
We perform cross validation by considering a random, individual splitting of the dataset between training and validation set.
After hyperparameter tuning, the final, optimal LFLDNet is trained on 1 Nvidia A40 GPU.
However, the code already accommodates data and model distributed training across multiple GPUs to further speed up computation for larger neural networks than those considered in this paper.

We do not perform gradient accumulation and we train LFLDNets in single-precision with the Adam optimizer \cite{Kingma2014} over 500 epochs for hyperparameter tuning and a maximum of 10.000 epochs for the final training, with an initial learning rate of 0.0003.
For all cases, we consider a batch size equal to 5.
In the training phase, at each epoch, 1.000 points are randomly sampled on the computational domain $\Omega$.
During inference, for GPU memory constraints, the mesh points $\spacevar \in \Omega$ are subdivided into 4 and 90 partitions for the electrophysiology and CFD test cases, respectively.
On each subdivision, the LFLDNet is queried over the entire time domain $t \in [0, T]$ to reproduce the whole solution field $\outputField(\spacevar, \timevar)$.
The number of subdivisions has been optimized to fulfill memory requirements according to the space-time resolution of the specific test case.
Although the dynamics network, focusing on the temporal behavior only, can be queried just once for all partitions, we consider the LFLDNet as a whole during inference.
This can certainly be optimized, but does not lead to a significant loss of performance, since the forward pass of the lightweight dynamics network is typically at least an order of magnitude faster than that of the reconstruction network.

For the Python implementation, we rely on the integration between PyTorch Lightning and the Ray distributed framework \cite{Ray2018} for training and inference of LFLDNets.
For the implementation of the dynamics network, we employ the publicly available \href{https://github.com/mlech26l/ncps}{NCPs GitHub repository} \cite{Lechner2020NCP}.

We monitor the Mean Square Error (MSE), that is:
\begin{equation}
\mathcal{L}(\outputField_\mathrm{obs}(\spacevar, \timevar), \outputField_\mathrm{pred}(\spacevar, \timevar)) = || \outputField_\mathrm{obs}(\spacevar, \timevar) - \outputField_\mathrm{pred}(\spacevar, \timevar) ||_\mathrm{L^2(\Omega; [0, T])}^2,
\end{equation}
where $\outputField_\mathrm{pred}(\spacevar, \timevar) = \ROMobs(\ROMrhsliquid(\inputSign(\timevar); \paramsROMdyn), [\cos(2 \pi \fourier \spacevar), \sin(2 \pi \fourier \spacevar)]^T, \spaceField(\spacevar), \spaceTimeField(\spacevar, t); \paramsROMobs)$.
We perform an adimensionalization of inputs and outputs in the $[-1, 1]$ interval.
Differently from \cite{Regazzoni2024}, we do not consider any $L^2$ regularization for $\ROMobs$ and $\ROMrhsliquid$ or other application-specific terms in the loss function.
Furthermore, the absence of an ODE-based solver for the dynamics network relaxes the need for a normalization time constant.

\subsubsection{Software and hardware}
\label{sec:methods:HPO}

All numerical simulations are performed with \texttt{svSolver} \cite{Updegrove2017} and \texttt{svFSIplus} \cite{Zhu2022}, multiphysics and multiscale finite element solvers for cardiovascular and cardiac modeling examples, respectively. Both solvers are available as part of the \texttt{SimVascular} suite of applications \cite{Updegrove2017}. Simulations were run on AMD-based and Intel-based Central Processing Unit (CPU) nodes available at the San Diego Super Computing Center (SDSC) Expanse cluster and Stanford Research Computing Center (SRCC), respectively.
A LFLDNet always runs on 1 Nvidia A40 GPU within the computational resources of SRCC.
Furthermore, this paper is accompanied by \url{https://github.com/StanfordCBCL/LFLDNets}, a GitHub repository containing a PyTorch Lightning-based code released under MIT License to run training/inference with LFLDNets and the dataset of numerical simulations collected for the different test cases.

\subsection{Cardiac electrophysiology}
\label{sec:methods:EP}

We introduce the mathematical model and the numerical scheme related to the first test case for cardiac electrophysiology.

\subsubsection{Mathematical model}
\label{sec:methods:EP:model}

We model cardiac electrophysiology in a 3D domain $\Omega = \Omega_\mathrm{BiV} \cup \Omega_\mathrm{purk} \subset \mathbb{R}^3$ represented by a biventricular model of a 7-year-old female pediatric patient affected by hypoplastic left heart syndrome using the monodomain equation \cite{Quarteroni2019,collifranzone2014book} coupled with the ten Tusscher-Panfilov ionic model \cite{TTP06}, that is:
\begin{eqnarray} \label{eqn:monodomain}
\left\{\begin{array}{ll}
\displaystyle
\frac{\partial \Pot}{\partial t}+\Iion(\Pot,\Ionic)
-\nabla\cdot(\DiffTens \nabla \Pot)=\Iapp(\spacevar, \timevar; \tLVstim) & \mbox{ in }\Omega\times(0,T],\\[2mm]
(\DiffTens\nabla \Pot)\cdot {\bf n}=0  & \mbox{ on }\partial\Omega\times(0,T],\\[2mm]
\displaystyle
\frac{d\Ionic}{dt}=\RhsIonic(\Pot,\Ionic; \GCaL, \GNa, \GKr)
& \mbox{ in }\Omega\times(0,T],\\[2mm]
\displaystyle
\Pot(\spacevar, 0) = \Pot_0(\spacevar), \
\Ionic(\spacevar, 0) = \Ionic_0(\spacevar) &  \mbox{ in } \Omega.
\end{array}\right.
\end{eqnarray}

The Purkinje system is generated by a fractal-tree method following prior work \cite{Costabal2016}. Transmembrane potential $\Pot$ describes the propagation of the electric signal over the cardiac tissue, vector $\Ionic=(y_1,\ldots,y_{M+P})$ defines the probability density functions of $M=12$ gating variables, which represent the fraction of open channels across the membrane of a single cardiomyocyte, and the concentrations of $P=6$ relevant ionic species.
Among them, intracellular calcium $Ca^{2+}$, sodium $Na^{+}$ and potassium $K^{+}$ play an important role for cardiac function \cite{Salvador2024DTCHD}.

The right hand side $\RhsIonic(\Pot,\Ionic)$ defines the dynamics of the gating and concentration variables.
The ionic current $\Iion(\Pot,\Ionic)$ models the effect of the cellular level ODEs on the organ scale PDE.
The analytical expressions of both $\RhsIonic(\Pot,\Ionic)$ and $\Iion(\Pot,\Ionic)$ derive from the mathematical formulation of the ten Tusscher-Panfilov ionic model \cite{TTP06}.

The diffusion tensor is expressed as $\DiffTens = \Di \IdentityVec + \Da \fZero \otimes \fZero$ in $\Omega_\mathrm{BiV}$ and $\DiffTens = \Dp \IdentityVec$ in $\Omega_\mathrm{purk}$, where $\fZero$ expresses the biventricular fiber field \cite{Piersanti2021}. $\Da, \Di, \Dp \in \mathbb{R}^+$ represent the anisotropic, isotropic and Purkinje conductivities, respectively.

We impose the condition of an electrically isolated domain by prescribing homogeneous Neumann boundary conditions $\partial\Omega$, where $\boldsymbol{n}$ is the outward unit normal vector to the boundary.

The action potential is triggered by a current $\Iapp({\bf x},t; \tLVstim)$ that is applied at time $t = 0$ at the tip of the right bundle branch, followed by another stimulus at a variable time $t = \tLVstim$ at the beginning of the left bundle branch. 

In Table~\ref{tab:parameterspace} we report descriptions, ranges and units for the 7 multiscale model parameters that we explore via Latin Hypercube Sampling to generate the dataset of 150 electrophysiology simulations (100 for training, 50 for validation).

\begin{table}[t!]
    \begin{center}
        \hspace*{-0.75cm}
        \begin{tabular}{l l l l}
            \toprule
            Parameter & Description & Range & Units \\
            \midrule
            $\GCaL$    & Maximal $Ca^{2+}$ current conductance & [1.99e-5, 7.96e-5]         & $\mathrm{cm}^3$ $\mathrm{ms}^{-1}$ $\mu \mathrm{F}^{-1}$ \\
            $\GNa$     & Maximal $Na^{+}$ current conductance  & [7.42, 29.68]              & $\mathrm{nS}$ $\mathrm{pF}^{-1}$ \\
            $\GKr$     & Maximal rapid delayed rectifier current conductance & [0.08, 0.31] & $\mathrm{nS}$ $\mathrm{pF}^{-1}$ \\
            $\Da$      & Anisotropic conductivity & [0.008298, 0.033192]                    & $\mathrm{mm}^{2}$ $\mathrm{ms}^{-1}$ \\
            $\Di$      & Isotropic conductivity & [0.002766, 0.011064]                      & $\mathrm{mm}^{2}$ $\mathrm{ms}^{-1}$ \\
            $\Dp$      & Purkinje conductivity & [1.0, 3.5]                                 & $\mathrm{mm}^{2}$ $\mathrm{ms}^{-1}$ \\
            $\tLVstim$ & Purkinje left bundle stimulation time & [0, 100]                   & $\mathrm{ms}$ \\
            \bottomrule
        \end{tabular}
        \caption{Parameter space sampled via latin hypercube for the cardiac electrophysiology test case.}
        \label{tab:parameterspace}
    \end{center}
\end{table}

\subsubsection{Numerical discretization}
\label{sec:methods:EP:discretization}

We perform space discretization of the monodomain equation coupled with the ten Tusscher-Panfilov ionic model using the Finite Element Method (FEM) with $\mathbb{P}_1$ Finite Elements.
The tetrahedral mesh of the biventricular-purkinje system is comprised of 1,016,192 cells and 240,555 DOFs.
We apply non-Gaussian quadrature rules to recover convergent conduction velocities of the biventricular-purkinje system \cite{Tikenogullari2023}.
Ionic conductances vary transmurally to account for epicardial, mid-myocardial and endocardial cells \cite{TTP06}.

For time discretization, we first update the variables of the ionic model and then the transmembrane potential by employing an Implicit-Explicit numerical scheme \cite{Regazzoni2022,Piersanti2022,Fedele2023}.
Specifically, in the monodomain equation, the diffusion term is treated implicitly and the ionic term is treated explicitly. Moreover, the ionic current is discretized by means of the Ionic Current Interpolation scheme \cite{Krishnamoorthi2013}. We employ a time step $\Delta t=0.1$ ms with Forward Euler scheme and we simulate one single heartbeat ($T = 0.6$ s).

We apply a Laplace-Dirichlet Rule-Based algorithm \cite{Bayer2012} to generate the fiber, sheet and sheet-normal distributions on the myocardium using the following parameters: $\alpha_\mathrm{epi}$ = $-60^\circ$, $\alpha_\mathrm{endo}$ = $60^\circ$, $\beta_\mathrm{epi}$ = $20^\circ$ and $\beta_\mathrm{endo}$ = $-20^\circ$.

\subsection{Cardiovascular Fluid Dynamics}
\label{sec:methods:CFD}

We present the mathematical and numerical details for the second test case in the framework of computational fluid dynamics (CFD).

\subsubsection{Mathematical model}
\label{sec:methods:CFD:model}

The Navier-Stokes equations model the flow of an incompressible Newtonian viscous fluid in a computational domain $\Omega = \Omega_\mathrm{aorta} \subset \mathbb{R}^3$, defined by a patient-specific aorta, in the time interval $(0, T)$.
The unknowns in primitive variables are velocity and pressure, i.e $(\boldsymbol{u}, p)$, and the mathematical formulation reads \cite{Pfaller2021}:

\begin{equation}
\begin{cases}
\rho \left( \dfrac{\partial \boldsymbol{u}}{\partial t} + \boldsymbol{u} \cdot \nabla \boldsymbol{u} \right) = - \nabla p + \mu \Delta \boldsymbol{u} & $in$ \; \Omega \times (0, T], \\
\nabla \cdot \boldsymbol{u} = 0 & $in$ \; \Omega \times (0, T], \\
\boldsymbol{u} = \boldsymbol{u}_0 & $in$ \; \Omega \times \{0\}, \\
Q(t) = \int_{\Gamma^\mathrm{inlet}} \boldsymbol{u}_\mathrm{inlet} \boldsymbol{n} d\Gamma = Q_\mathrm{inlet}(t) & $on$ \; \Gamma^\mathrm{inlet} \times (0, T], \\
\boldsymbol{u} = \boldsymbol{0} & $on$ \; \Gamma^\mathrm{wall} \times (0, T], \\
p(t) = f_{\mathcal{R}\mathcal{C}\mathcal{R}}(Q_\mathrm{outlet}^i(t), \dot{Q}_\mathrm{outlet}^i(t), R_p^i, C^i, R_d^i) & $on$ \; \Gamma^\mathrm{outlet, i} \times (0, T], \; i = 1, ..., 5,
\label{eqn: NavierStokes}
\end{cases}
\end{equation}
where $\rho = 1.06 \, \mathrm{g} \, \mathrm{cm}^{-3}$ the blood density and $\mu = 0.04 \, \mathrm{g} \, \mathrm{cm}^{-1} \, \mathrm{s}^{-1}$ the dynamic viscosity.
The first equation represents momentum conservation, whereas the second governs mass conservation.
We apply a time-dependent, pulsatile inflow profile at the aortic inlet $\Gamma^\mathrm{inlet}$, no-slip Dirichlet boundary conditions on the outer wall $\Gamma^\mathrm{wall}$, and three-element Windkessel-type boundary conditions, also known as RCR models, at the five different outlets $\Gamma^\mathrm{outlet, i} \; i = 1, ..., 5$ where, for each outlet, we consider a proximal resistance $R_p$ in series with a parallel distal resistance $R_d$ and capacitance $C$ \cite{Taylor2010}.
For further mathematical details we refer to \cite{Updegrove2017}.

\begin{table}[t!]
    \begin{center}
        \begin{tabular}{l l l l}
            \toprule
            Parameter & Description & Range & Units \\
            \midrule
            $R_p^i$               & Proximal resistance in RCR boundary conditions & [85.1, 1293.5] & $\mathrm{Pa}$ $\mathrm{s}$ $\mathrm{mL}^{-1}$ \\
            $C^i$                 & Capacitance in RCR boundary conditions  & [6.9e-5, 8.2e-4]    & $\mathrm{mL}$ $\mathrm{Pa}^{-1}$ \\
            $R_d^i$               & Distal resistance in RCR boundary conditions & [126.9, 17442.9] & $\mathrm{Pa}$ $\mathrm{s}$ $\mathrm{mL}^{-1}$ \\
            $Q_\mathrm{inlet}(t)$ & Inlet volumetric flow rate & [0, 448]                    & $\mathrm{mL}$ $\mathrm{s}^{-1}$ \\
            \bottomrule
        \end{tabular}
        \caption{Parameter space of the cardiovascular CFD test case. Note that the ranges for the RCR boundary conditions include all five outlets (i.e., $i = 1, ..., 5$), which can start from very different baseline values.}
        \label{tab:CFD:parameterspace}
    \end{center}
\end{table}

In Table~\ref{tab:CFD:parameterspace} we report descriptions, ranges and units for the inflow and outflow boundary conditions that we sample to generate the dataset of 32 CFD simulations (25 for training, 7 for validation).
Following \cite{Pegolotti2024}, we multiplied a baseline inlet flow rate and each of the parameters governing the outlet boundary conditions by independent factors uniformly distributed in the range [0.8, 1.2].

\subsubsection{Numerical discretization}
\label{sec:methods:CFD:discretization}

We perform space discretization of the Navier-Stokes equations using FEM with a Streamline-Upwind/Petrov-Galerkin (SUPG) and Pressure-Stabilizing/Petrov-Galerkin (PSPG) $\mathbb{P}_1$-$\mathbb{P}_1$ formulation \cite{Whiting2001}. The patient-specific aorta, which is taken from the Vascular Model Repository \cite{Wilson2013}, is comprised of 3,018,813 cells and 560,868 DOFs.

This stabilized FEM formulation is discretized in time using the generalized alpha time stepping scheme \cite{Liu2021} with a time step size $\Delta t = 0.001$ s.
We run the CFD simulation for 5 cardiac cycles, with a period that is equal to $0.85$ s, and we retain the last one for training and inference.
For further details regarding numerical discretization within the SimVascular solver we refer to \cite{Updegrove2017}.






	\section{Results}
\label{sec:results}

We build and present two surrogate models of cardiac and cardiovascular function using LFLDNets that effectively learn the parameterized spatio-temporal dynamics underlying sets of multiscale and multiphysics PDEs arising in computational electrophysiology and fluid dynamics.

\subsection{Test case 1: cardiac electrophysiology}
\label{sec:results:EP}

\begin{table}[h!]
    \centering
    \begin{tabular}{c cccc}
        \toprule
        & \multicolumn{4}{c}{Hyperparameters} \\
        & $\ROMrhsliquid$/$\ROMobs$ neurons & $\ROMobs$ layers & $N_f$ & $N_s$ \\
        \midrule
        tuning & $\{200, 250, \; ... \;, 400\}$ & $\{5, 10, 15\}$ & $\{25, 50, \; ... \;, 200\}$ & $\{ 50, 100, 150 \}$ \\
        final    & 200 & 10 & 50 & 50 \\
        \bottomrule
    \end{tabular}
    \caption{LFLDNets hyperparameter tuning for the electrophysiology test case. Hyperparameter ranges (top) and optimized values (bottom). The final model sizes are $\ROMrhsliquid$ (283K parameters), $\ROMobs$ (432K parameters) and $\fourier$ (150 parameters).}
    \label{tab:HPO_EP}
\end{table}

\begin{figure}
     \centering
     \begin{subfigure}[b]{0.46\textwidth}
         \centering
         \includegraphics[width=\textwidth]{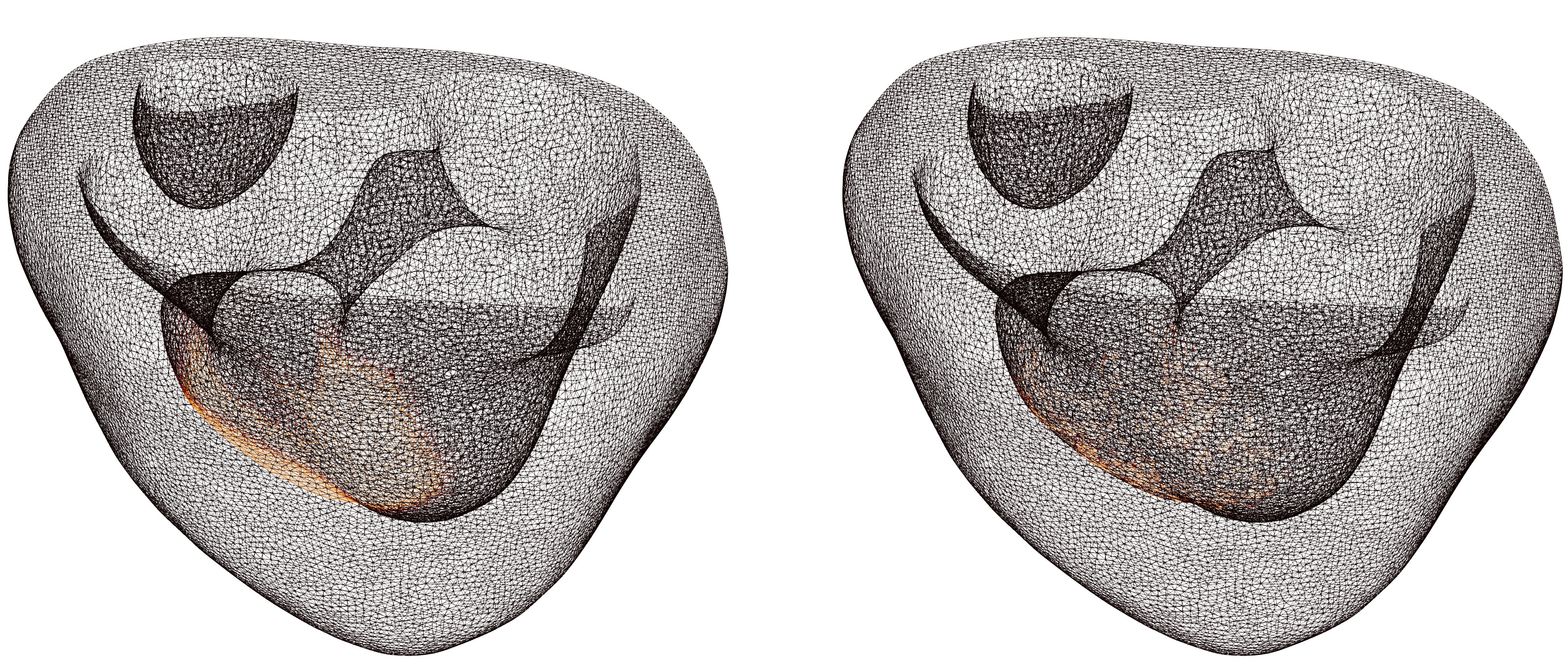}
         \caption{$t = 20$ ms}
         \label{fig:20}
     \end{subfigure}
     \hfill
     \begin{subfigure}[b]{0.46\textwidth}
         \centering
         \includegraphics[width=\textwidth]{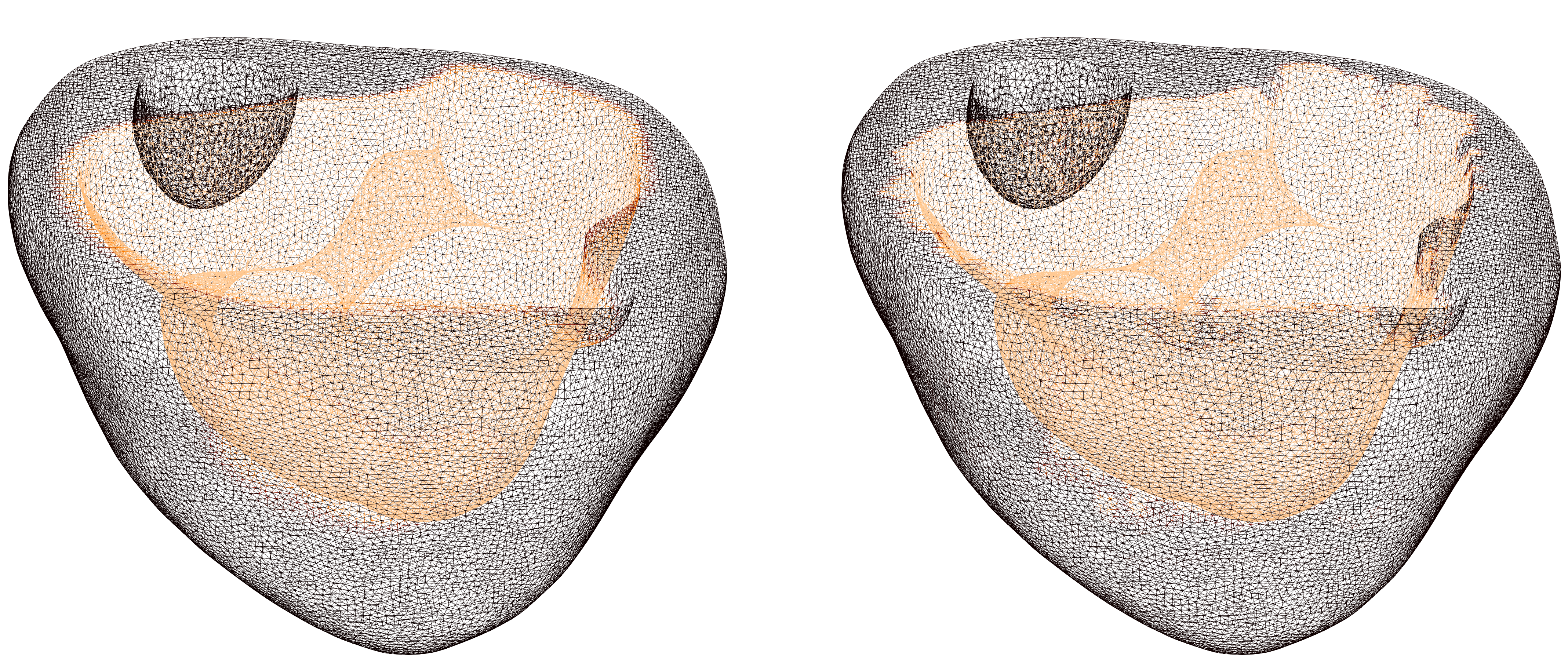}
         \caption{$t = 50$ ms}
         \label{fig:50}
     \end{subfigure}
     \hfill
     \begin{subfigure}[b]{0.46\textwidth}
         \centering
         \includegraphics[width=\textwidth]{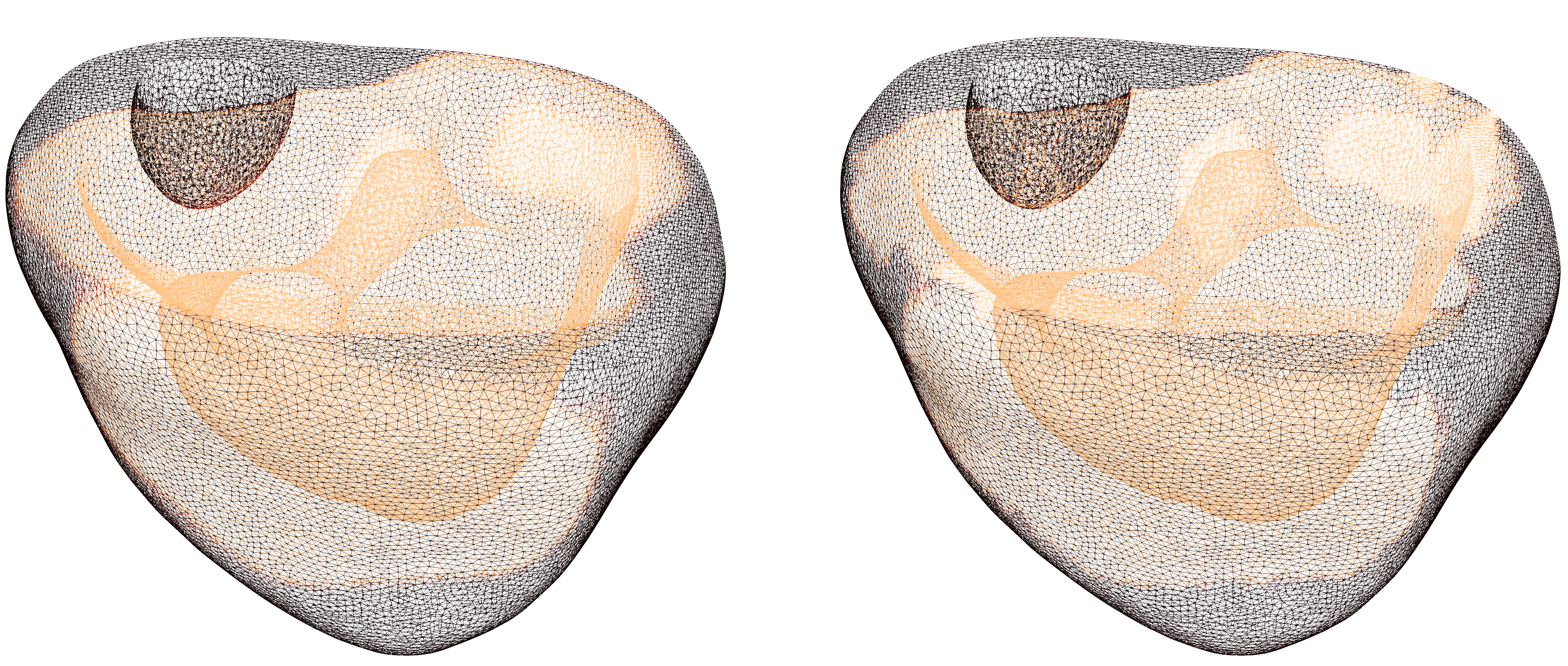}
         \caption{$t = 60$ ms}
         \label{fig:60}
     \end{subfigure}
     \hfill
     \begin{subfigure}[b]{0.46\textwidth}
         \centering
         \includegraphics[width=\textwidth]{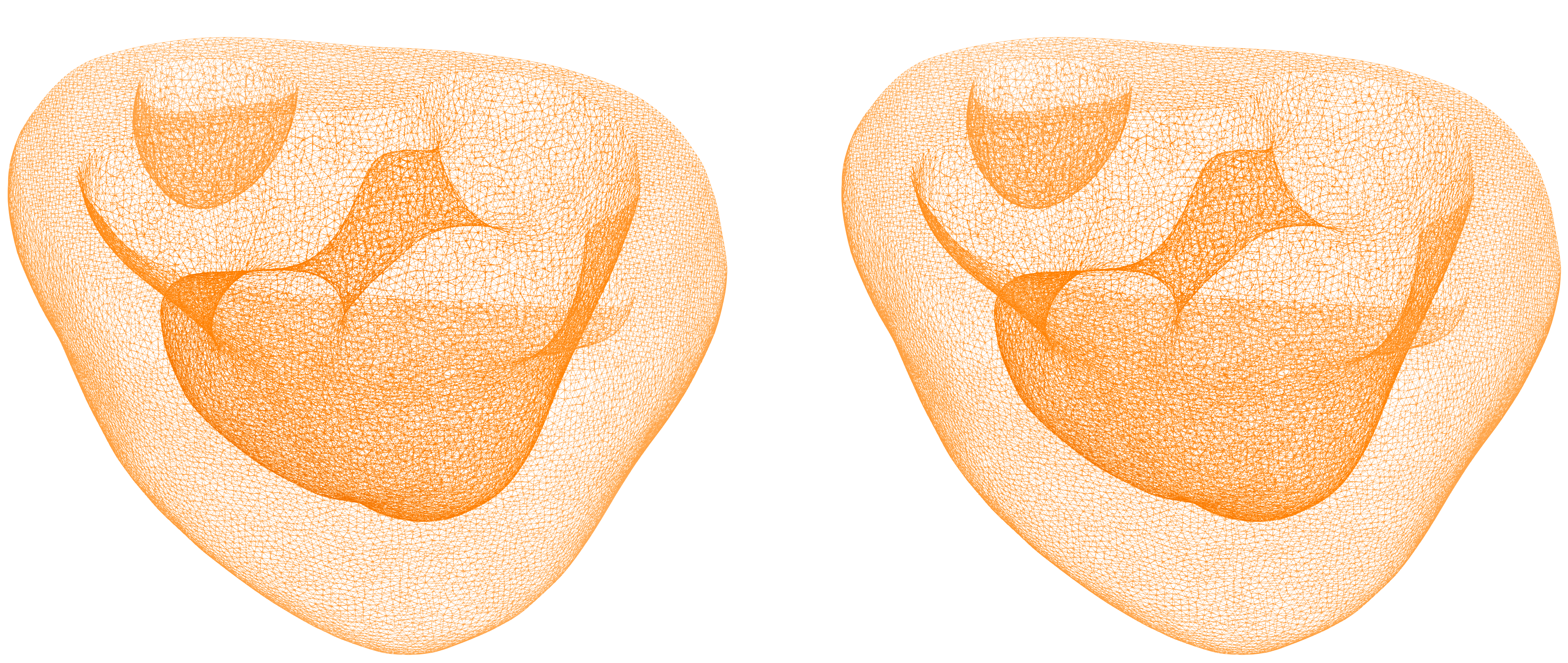}
         \caption{$t = 200$ ms}
         \label{fig:210}
     \end{subfigure}
     \hfill
     \begin{subfigure}[b]{0.46\textwidth}
         \centering
         \includegraphics[width=\textwidth]{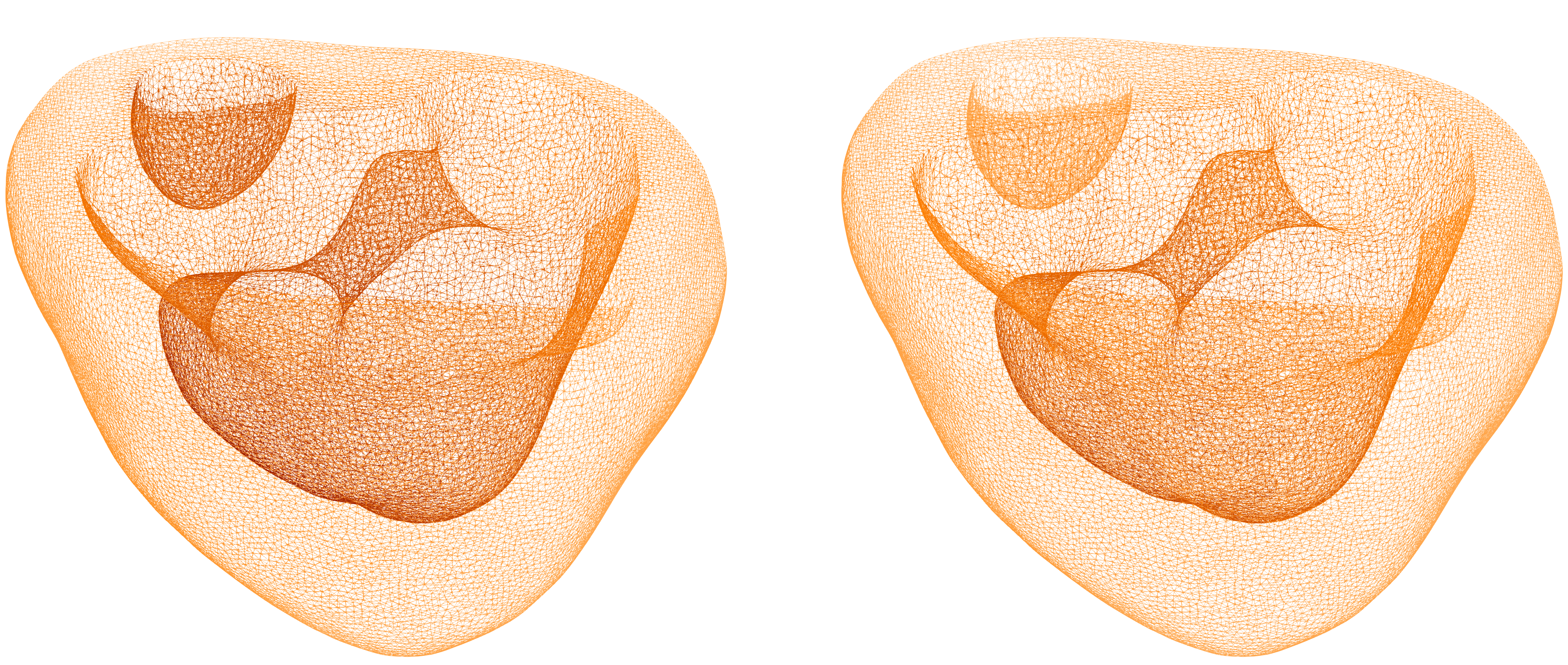}
         \caption{$t = 250$ ms}
         \label{fig:260}
     \end{subfigure}
     \hfill
     \begin{subfigure}[b]{0.46\textwidth}
         \centering
         \includegraphics[width=\textwidth]{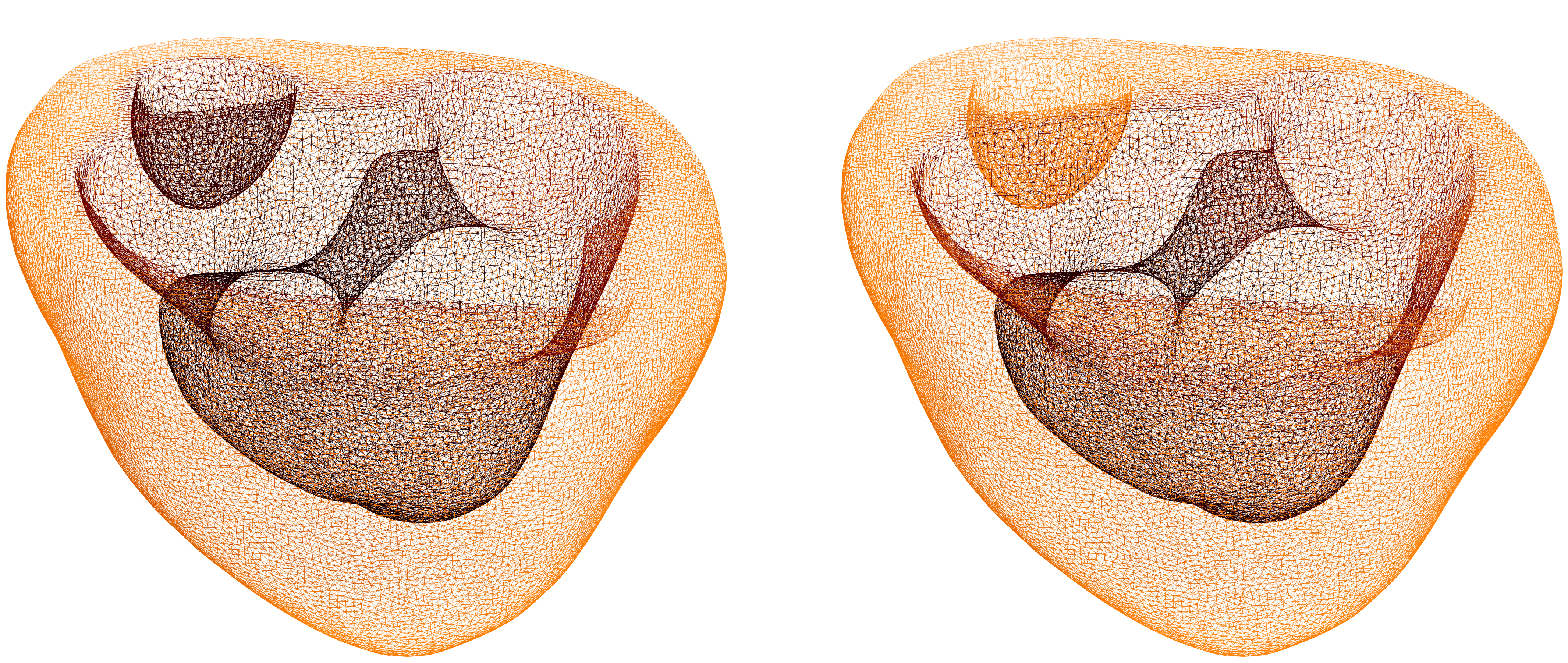}
         \caption{$t = 290$ ms}
         \label{fig:290}
     \end{subfigure}
     \hfill
     \begin{subfigure}[b]{0.49\textwidth}
         \centering
         \includegraphics[width=\textwidth]{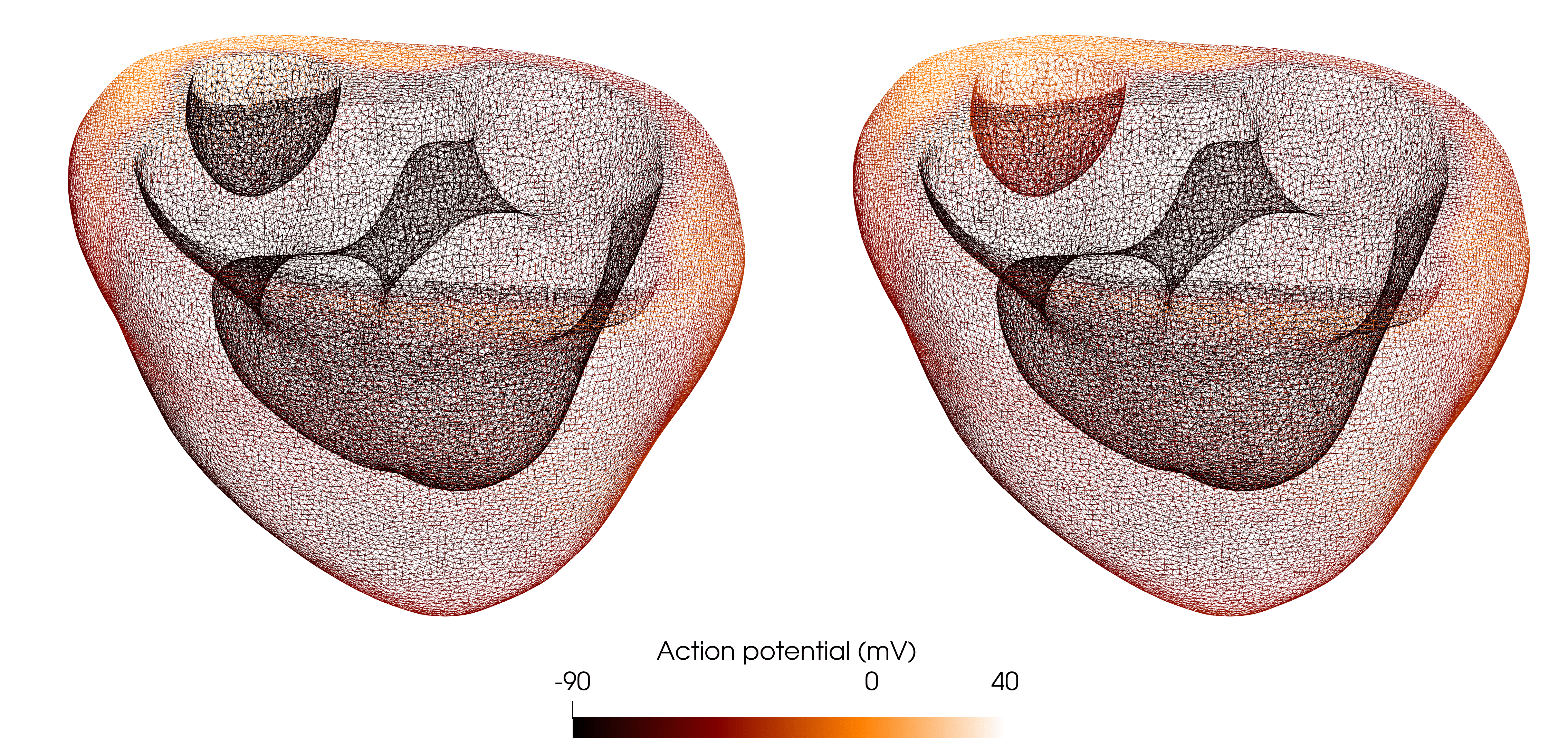}
         \caption{$t = 330$ ms}
         \label{fig:330}
     \end{subfigure}
     \caption{Electrophysiology test case. Comparison of the action potential $u(\spacevar, \timevar)$ throughout the heartbeat for a random sample in the validation set. LFLDNet prediction (left), ground truth from electrophysiology simulation (right).}
     \label{fig:HLHS_actionpotential}
\end{figure}

\begin{figure}
     \centering
     \begin{subfigure}[b]{0.25\textwidth}
         \centering
         \includegraphics[width=\textwidth]{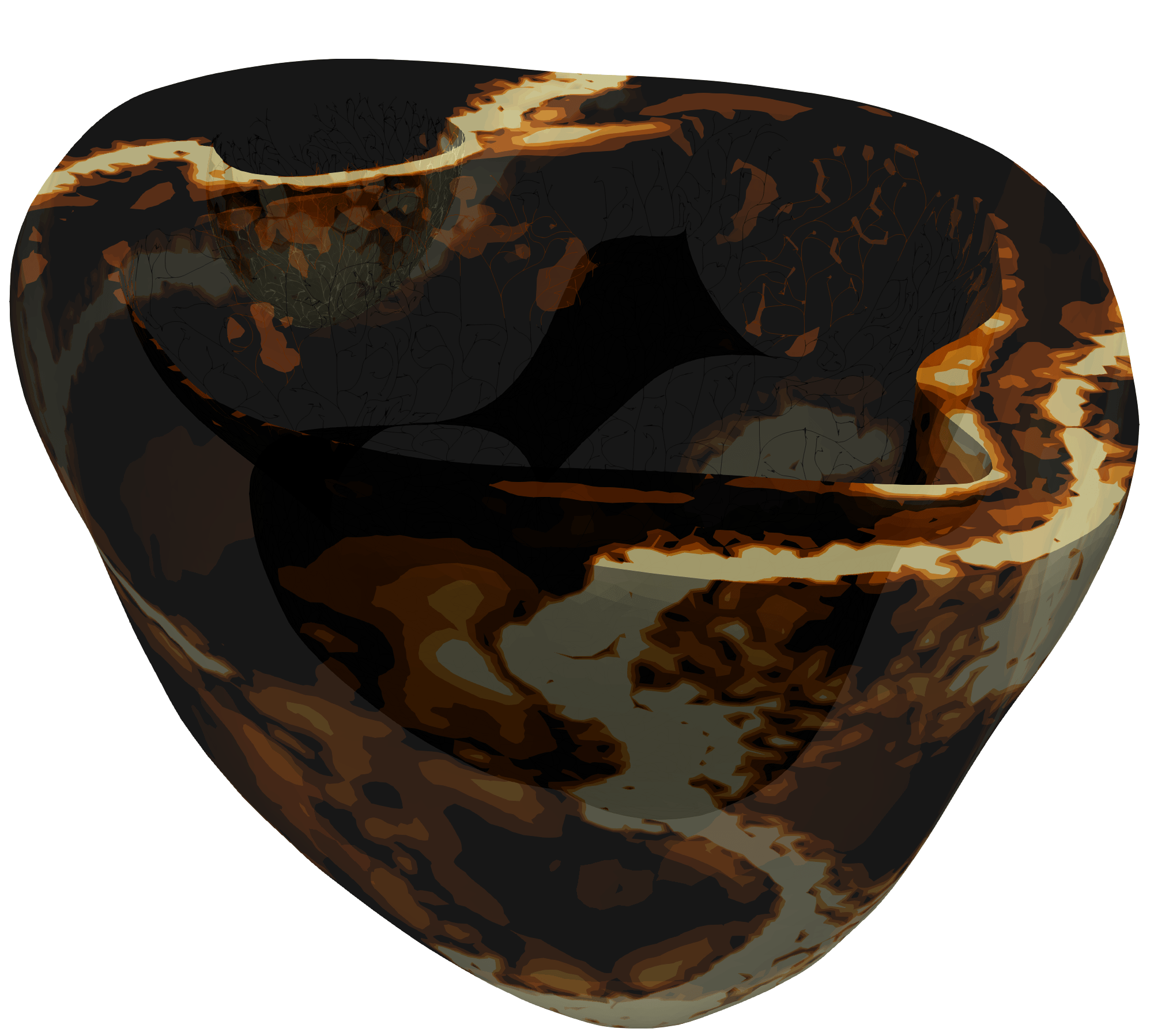}
         \caption{$t = 60$ ms}
         \label{fig:60}
     \end{subfigure}
     \hfill
     \begin{subfigure}[b]{0.25\textwidth}
         \centering
         \includegraphics[width=\textwidth]{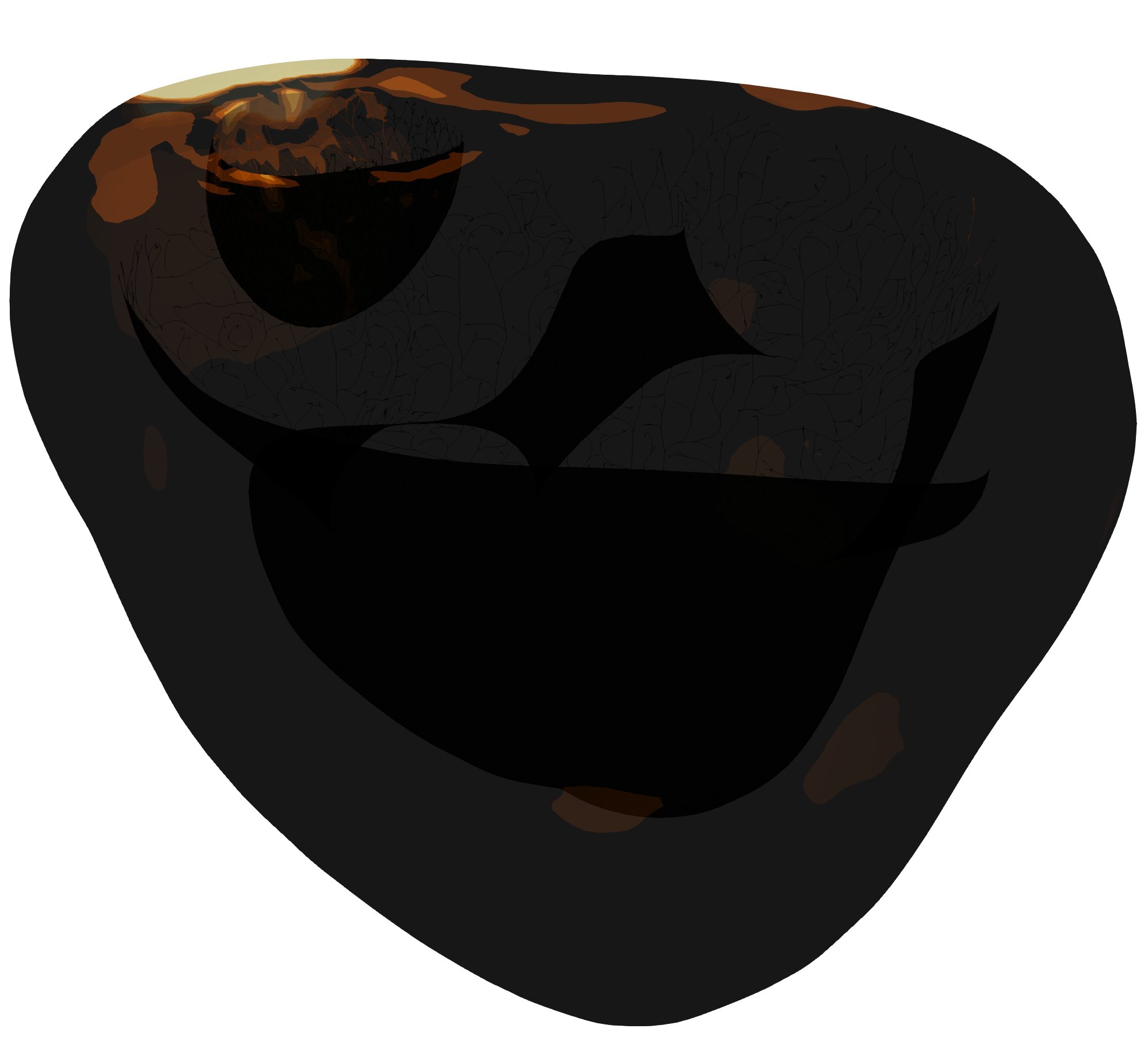}
         \caption{$t = 70$ ms}
         \label{fig:70}
     \end{subfigure}
     \hfill
     \begin{subfigure}[b]{0.25\textwidth}
         \centering
         \includegraphics[width=\textwidth]{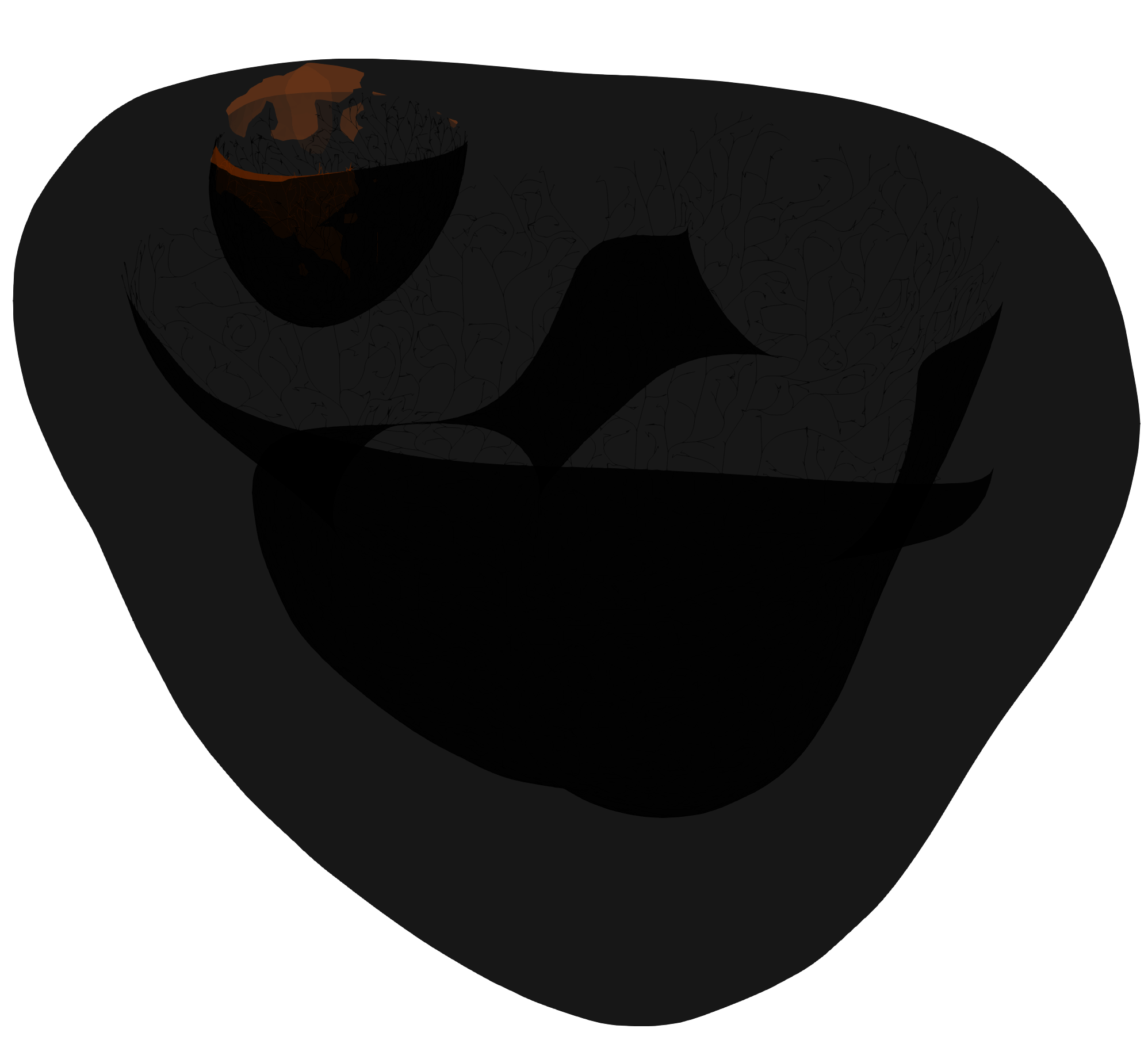}
         \caption{$t = 100$ ms}
         \label{fig:100}
     \end{subfigure}
     \hfill
     \begin{subfigure}[b]{0.25\textwidth}
         \centering
         \includegraphics[width=\textwidth]{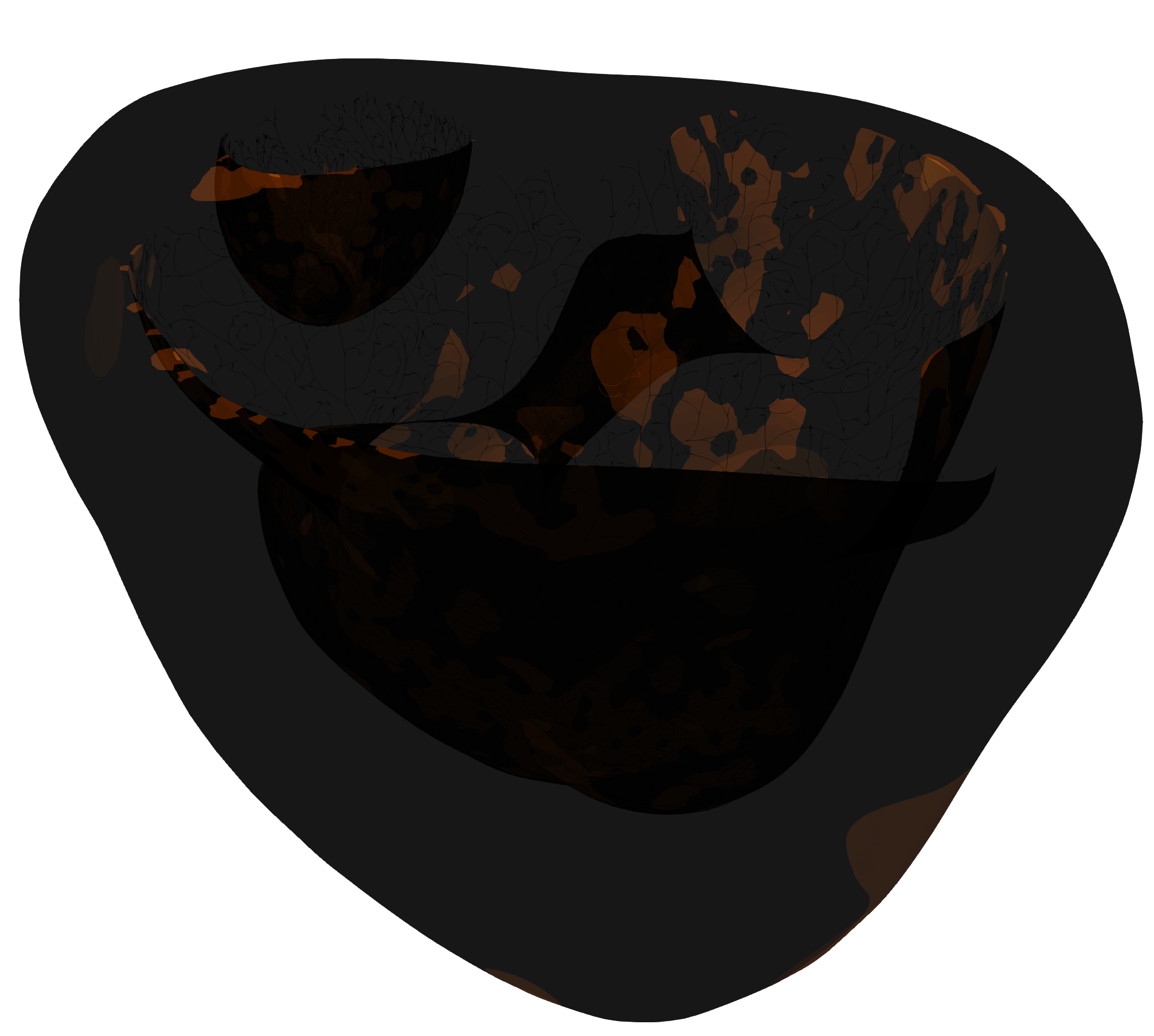}
         \caption{$t = 200$ ms}
         \label{fig:200}
     \end{subfigure}
     \hfill
     \begin{subfigure}[b]{0.25\textwidth}
         \centering
         \includegraphics[width=\textwidth]{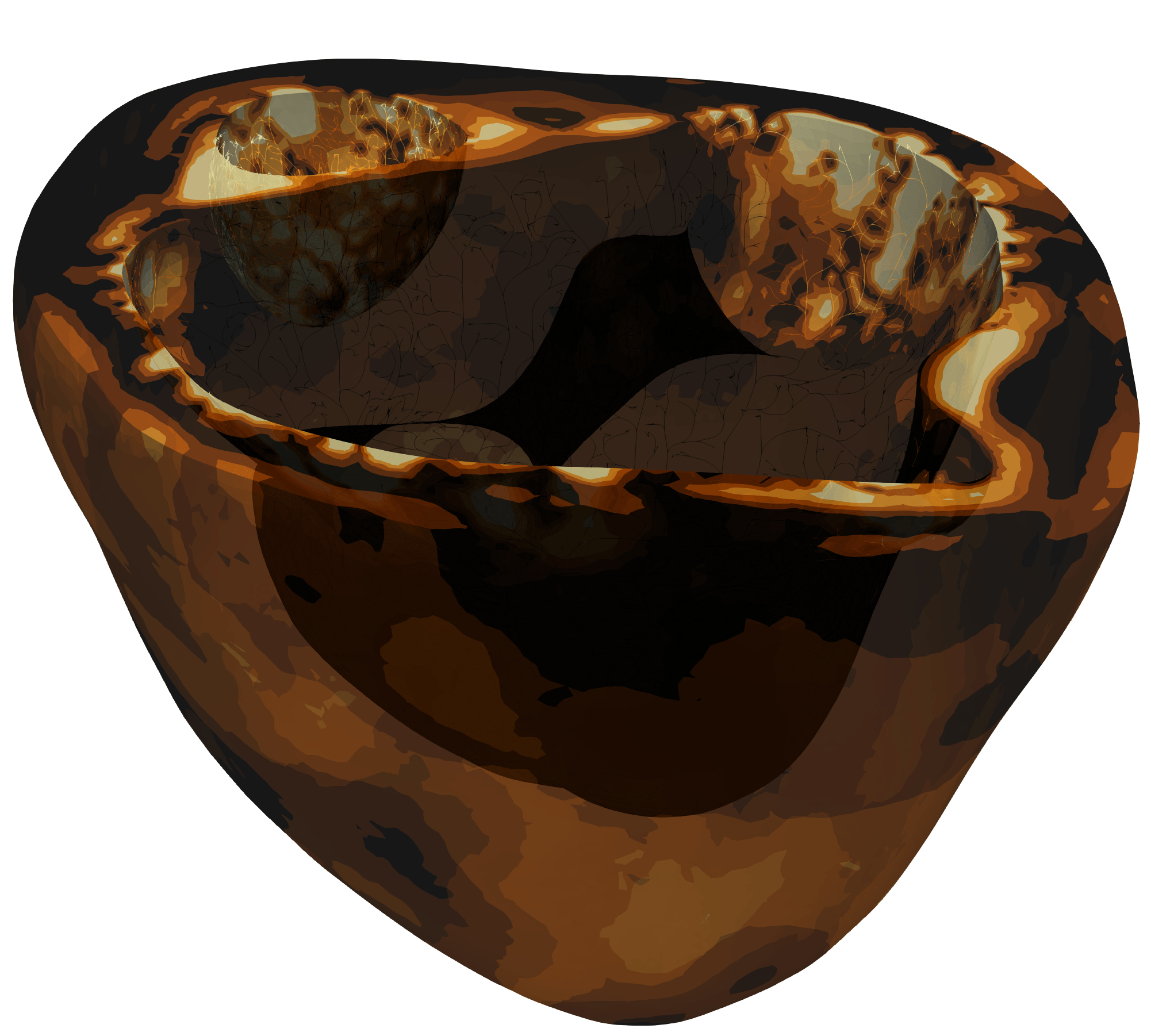}
         \caption{$t = 300$ ms}
         \label{fig:300}
     \end{subfigure}
     \hfill
     \begin{subfigure}[b]{0.25\textwidth}
         \centering
         \includegraphics[width=\textwidth]{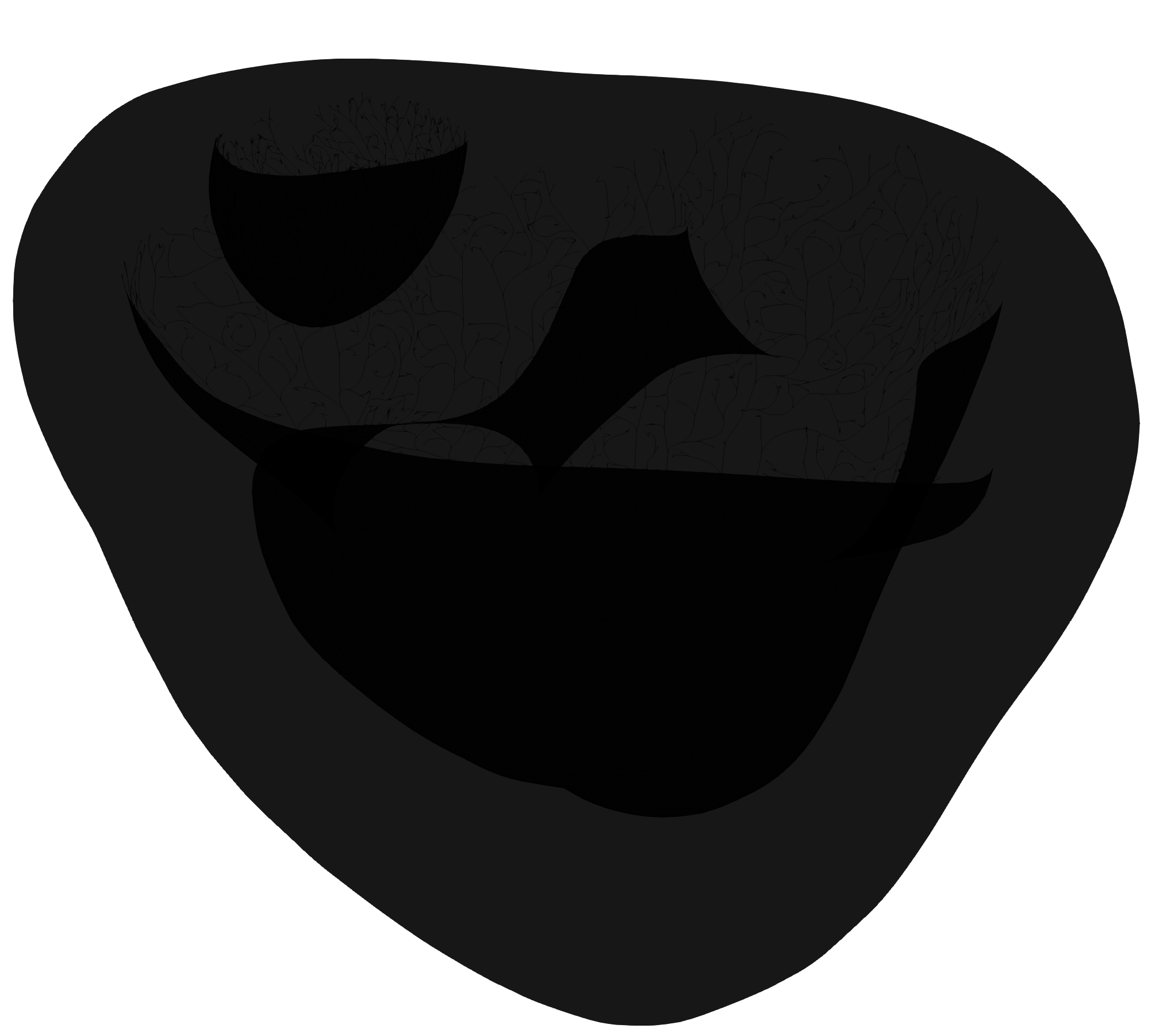}
         \caption{$t = 400$ ms}
         \label{fig:400}
     \end{subfigure}
     \hfill
     \begin{subfigure}[b]{0.25\textwidth}
         \centering
         \includegraphics[width=\textwidth]{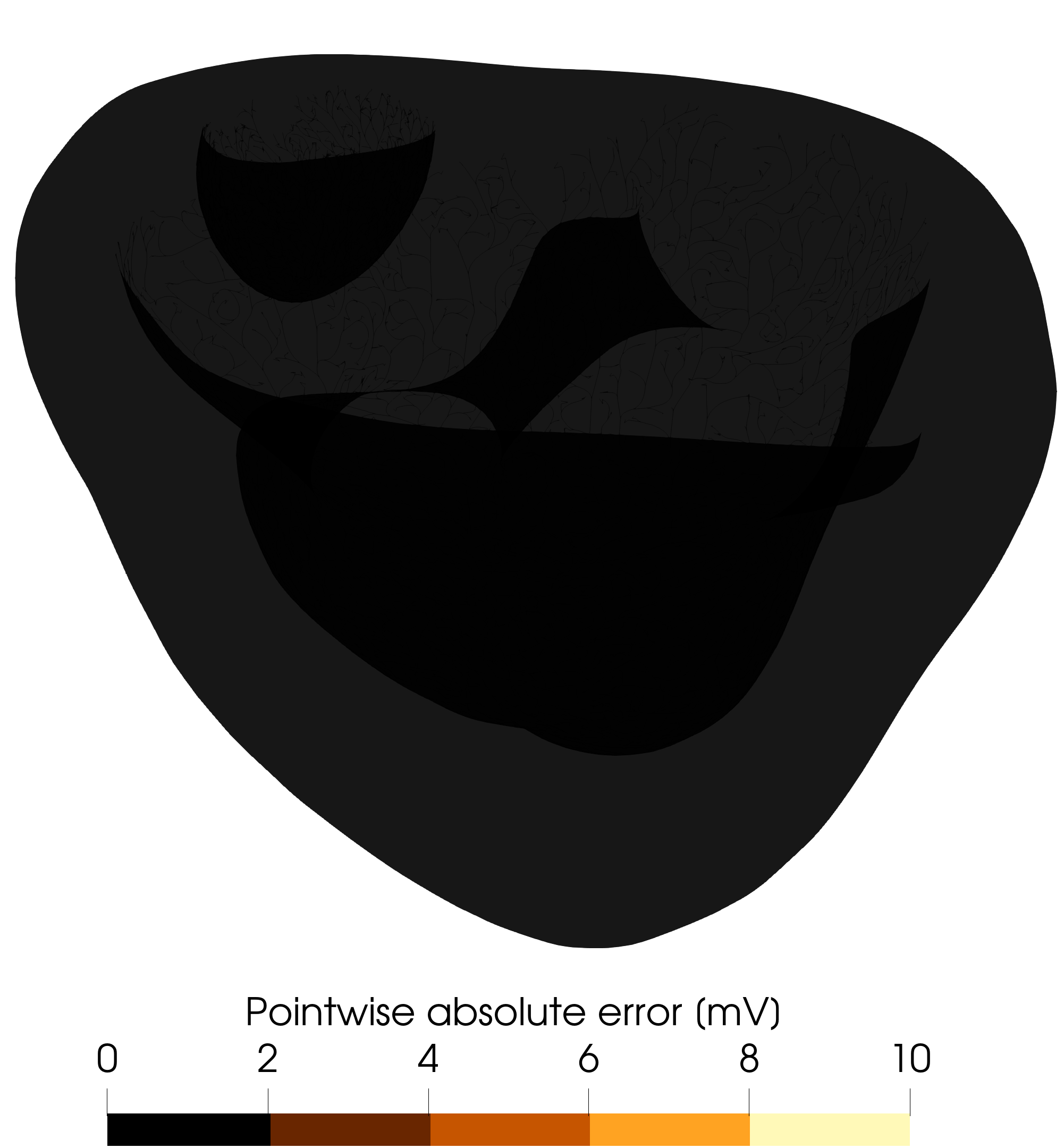}
         \caption{$t = 500$ ms}
         \label{fig:500}
     \end{subfigure}
     \caption{Electrophysiology test case. Pointwise absolute difference $|u_\mathrm{pred}(\spacevar, \timevar) - u_\mathrm{obs}(\spacevar, \timevar)|$ between LFLDNets predictions and observations over the cardiac cycle for a random sample in the validation set.}
     \label{fig:HLHS_errors}
\end{figure}

\begin{figure}[t!]
    \centering
    \includegraphics[width=1.0\textwidth]{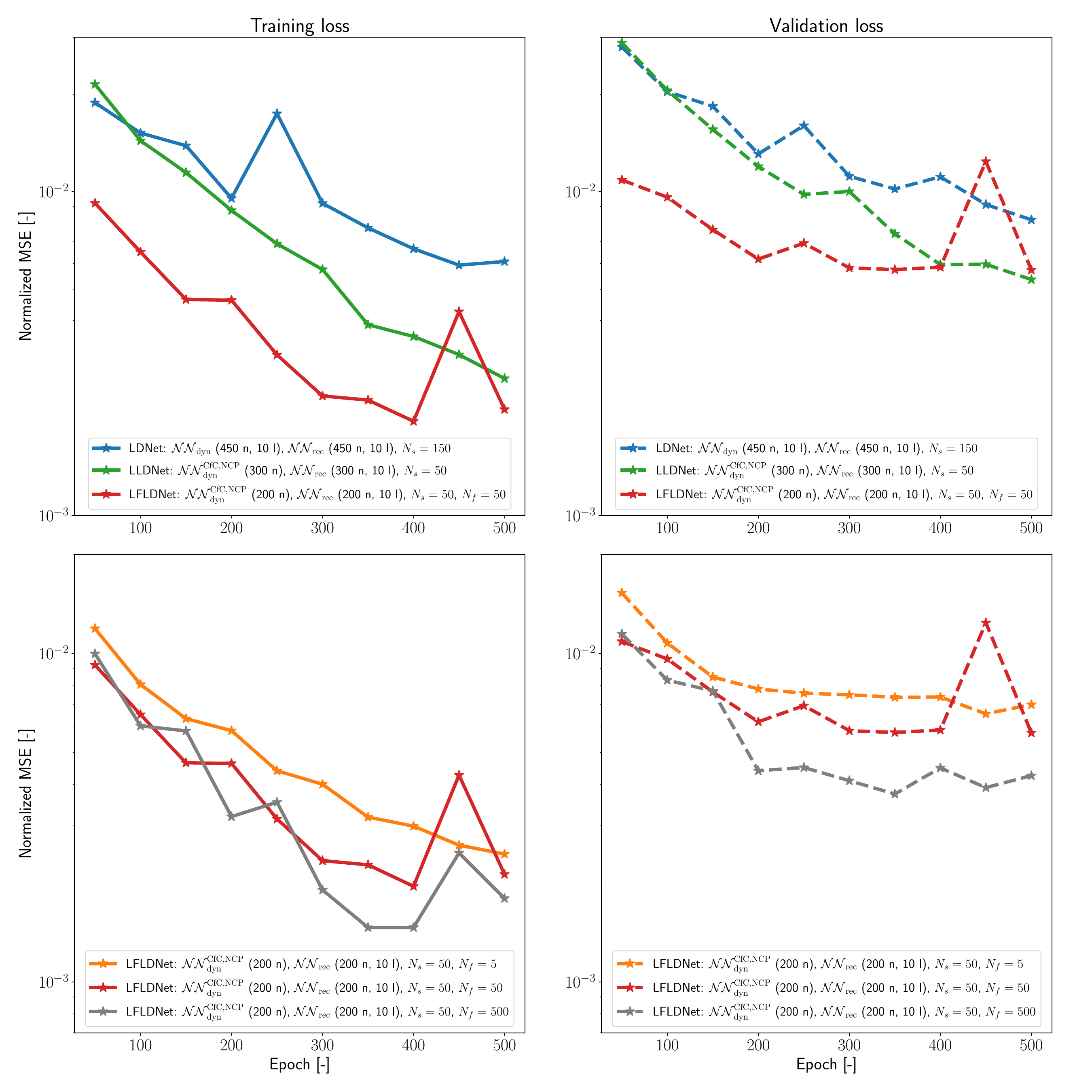}
    \caption{Electrophysiology test case. Comparison of training and validation losses between LDNets, LLDNets, and LFLDNets after hyperparameter tuning (top). Comparison of training and validation losses for different Fourier embedding dimensions within the optimal LFLDNets (bottom).}
    \label{fig:HLHS_comparisons}
\end{figure}

\begin{figure}[t!]
    \centering
    \subfloat[\centering LDNets]{{\includegraphics[width=8cm]{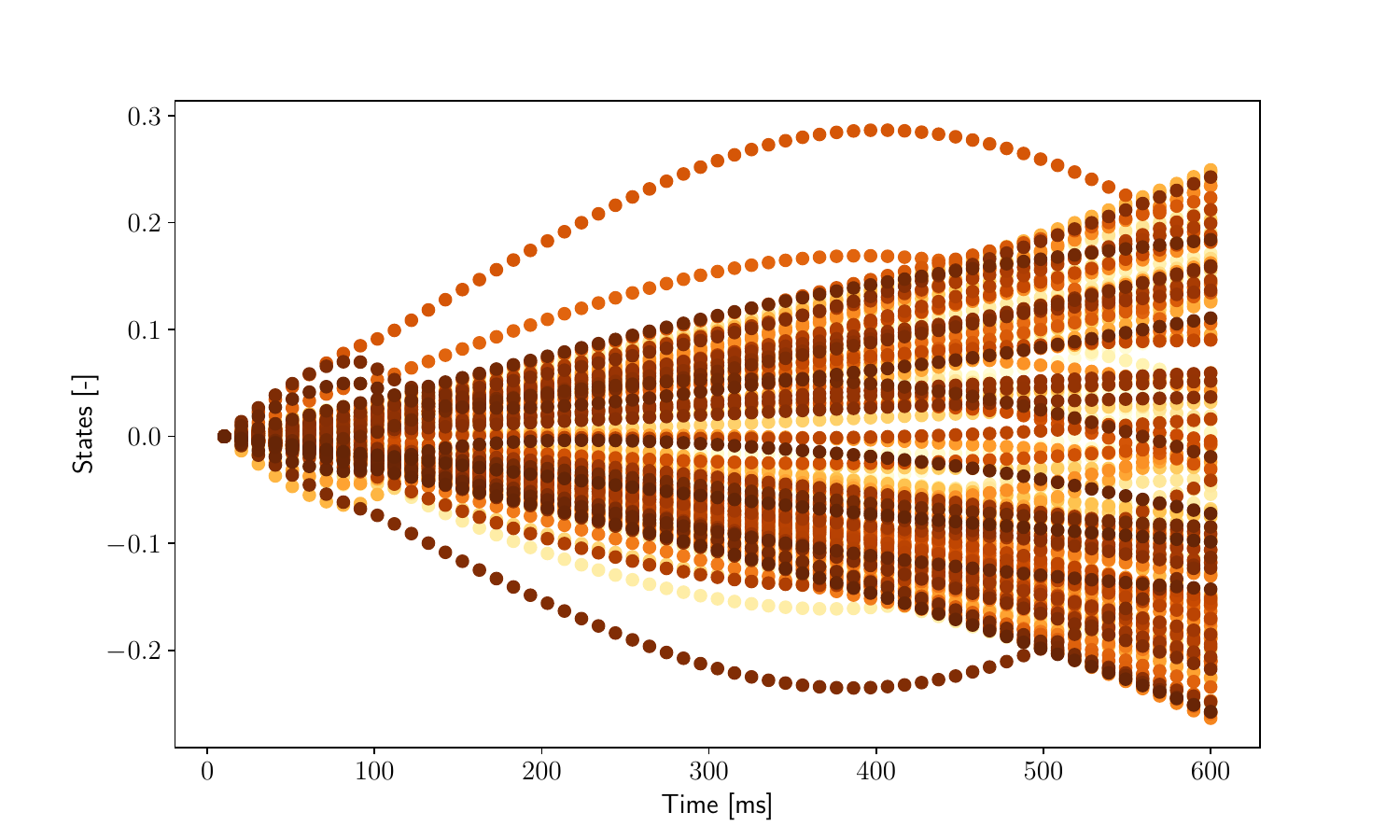}}}
    \qquad
    \subfloat[\centering LLDNets]{{\includegraphics[width=8cm]{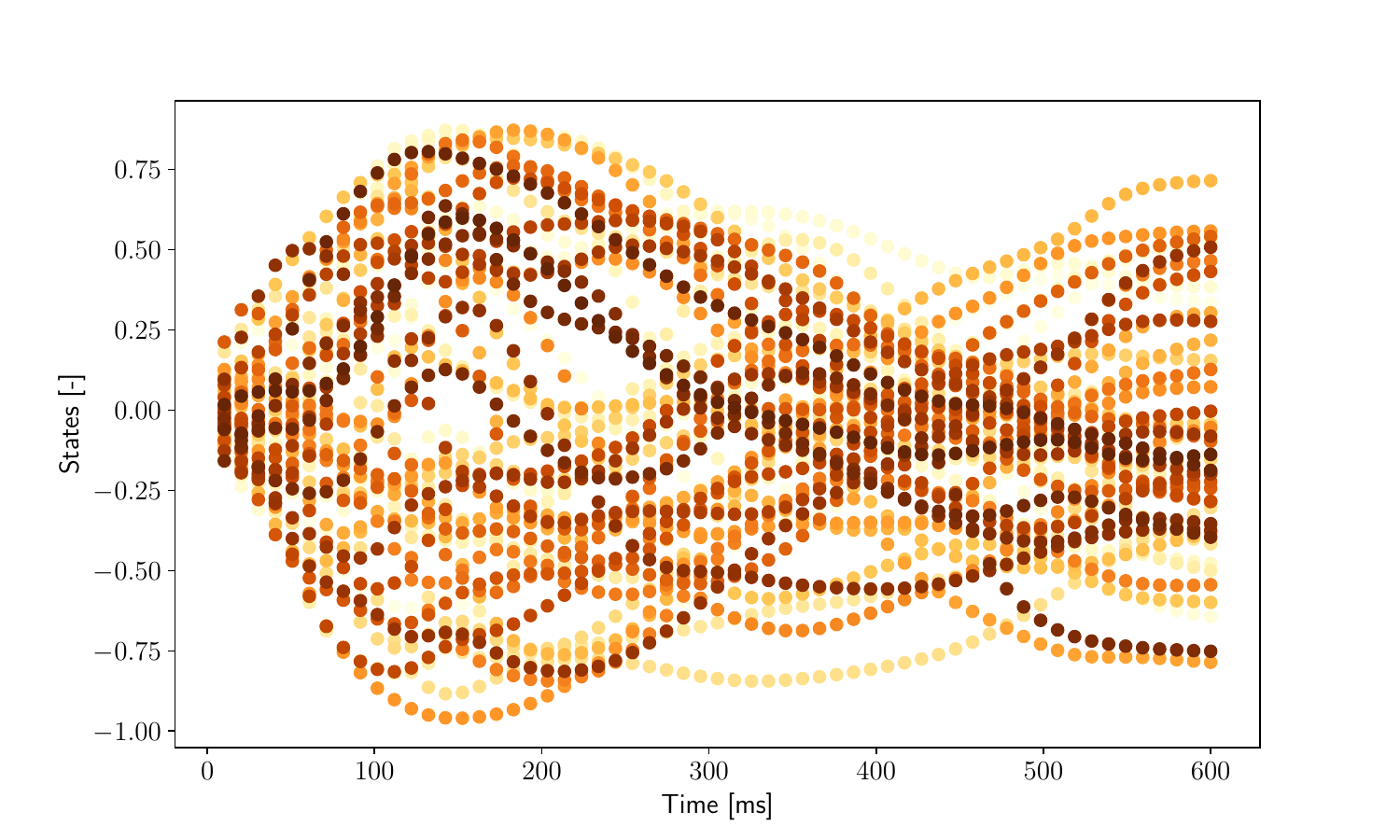}}}
    \qquad
    \subfloat[\centering LFLDNets]{{\includegraphics[width=8cm]{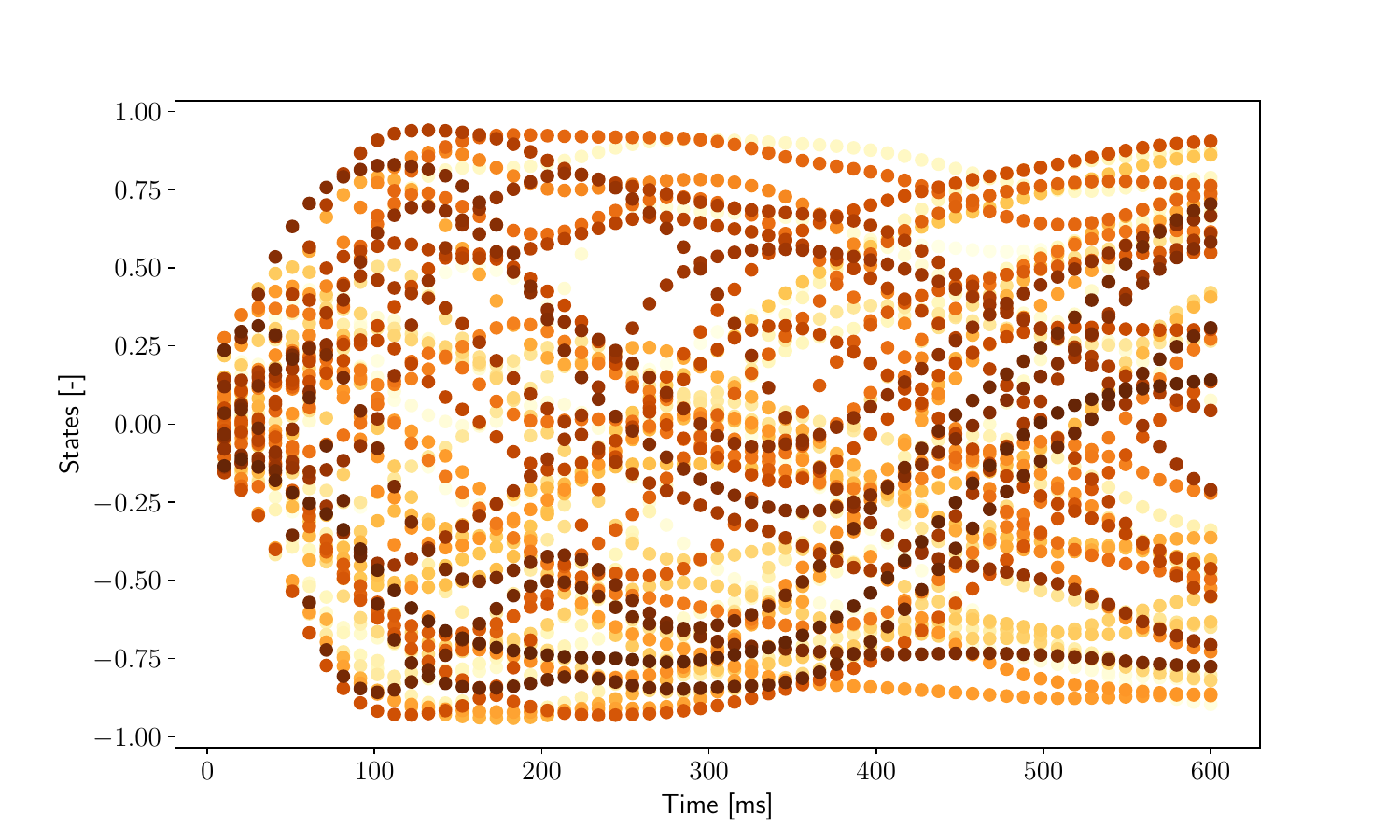}}}
    \caption{Electrophysiology test case. Evolution of the global state vector $\stateROM(\timevar)$ over the cardiac cycle for a specific choice of the input signals $\inputSign(\timevar)$ coming from a random sample in the validation set. Different colors identify different components of $\stateROM(\timevar)$. Optimal LDNet (top), optimal LLDNet (center) and optimal LFLDNet (bottom).}
    \label{fig:HLHS_states}
\end{figure}

In Table~\ref{tab:HPO_EP} we report the range of hyperparameters we consider for tuning, along with the final set chosen by the Bayesian optimizer.
We see that the optimal LFLDNet is compact, with less than a million tunable parameters.
This is remarkable given the complexity of the 3D multiscale electrophysiology simulations encoded by this surrogate model.

Training LFLDNets for hyperparameter tuning over 500 epochs takes from 10 minutes to 5 hours, depending on the specific architecture.
The optimal configuration achieves convergence at epoch 1.799 after 1 hour and 30 minutes of computation.
Specifically, the normalized MSEs for the training (100 simulations) and validation (50 simulations) sets are 9.12e-4 and 3.15e-3, respectively, at epoch 1.799.
Running inference over the whole spatio-temporal numerical simulation requires 3 minutes in eager mode.
On the other hand, a high-fidelity simulation takes about 1.5 hours on 24 cores.
In Figures~\ref{fig:HLHS_actionpotential} and~\ref{fig:HLHS_errors} we depict the spatio-temporal evolution of the action potential and the pointwise absolute errors over the cardiac cycle at different time points for a random sample of the validation set, respectively.
We show good agreement between the LFLDNets prediction and the original numerical simulation throughout the heartbeat, especially with respect to the overall propagation pattern and conduction velocities.
The pointwise absolute error remains uniformly small over the cardiac cycle and increases only locally in correspondence with the wavefront during some instances of the depolarization and repolarization phases, which exhibit steep gradients.
We also note that the uniform time step chosen for training and inference, i.e. $\Delta t=10$ ms is orders of magnitude larger than the maximum time step allowed in the Finite Element simulations of the monodomain equation coupled with the ten Tusscher-Panfilov ionic model, due to stiffness, convergence and stability requirements.
Furthermore, this test case presents bifurcations in the parameter space, as the simulated behavior ranges from sinus rhythm, which is a healthy propagation pattern, to different stages of bundle branch block, which is a pathological condition.
As mentioned in the Methods section, all LFLDNets operations are performed on 1 Nvidia A40 GPU.

In Figure~\ref{fig:HLHS_comparisons} we compare the performance of LFLDNets with their original LDNet and LLDNet counterparts, i.e. LFLDNets without the Fourier embedding.
We perform hyperparameter tuning for both LDNets and LLDNets using the same ranges of Table~\ref{tab:HPO_EP}.
We see that both architectures require larger neural networks than the optimal LFLDNet, which consists of 200 neurons for the dynamics network, 10 hidden layers with 200 neurons per layer for the reconstruction network.
Specifically, the optimal LDNet has 10 hidden layers with 450 neurons per layer for both networks, whereas LLDNet presents 300 neurons for the dynamics network, 10 hidden layers with 300 neurons per layer for the reconstruction network.
We remark that CfC, NCPs have a fixed number of hidden layers, which is set to 4, and the only tunable hyperparameter is given by the total amount of neurons, which is divided into sensory, inter, command, and motor neurons according to the specified number of inputs and states.
We also notice that the optimal dimension of the state vector is 150 for LDNet and 50 for LLDNet/LFLDNet, resulting in a smaller state representation for the liquid counterpart.
Furthermore, the lighter LFLDNet manifests faster convergence and lower normalized MSEs from the very first epochs.
Training the optimal LDNet, LLDNet and LFLDNet for 500 epochs takes approximately 30, 40 and 50 minutes, respectively.

In Figure~\ref{fig:HLHS_comparisons} we also perform a convergence study in terms of the dimension of the Fourier features, where we show that larger embeddings are associated with better convergence to smaller normalized MSEs.
However, this comes at the cost of more expensive training and slightly higher memory requirements to store a larger Fourier embedding.

In Figure~\ref{fig:HLHS_states} we compare LDNets, LLDNets and LFLDNets in terms of the state vector $\stateROM(\timevar)$ temporal dynamics for a random sample of the validation set.
We see that 150 LDNets states mostly exhibit stretched and compact trajectories, whereas 50 states for both LLDNets and LFLDNets show smooth trajectories that are varying throughout the heartbeat according to the features of the cardiac electrophysiology simulation and cover more uniformly the bounded [-1, 1] non-dimensional range.
This motivates why the optimal architectures for LLDNets and LFLDNets have a lower number of states than LDNets.

Overall, we show that LFLDNets demonstrate superior performance in terms of number of tunable parameters for the dynamics and reconstruction networks, generalization errors, state vector temporal dynamics, training and inference times with respect to LDNets.

\subsection{Test case 2: cardiovascular fluid dynamics}
\label{sec:results:CFD}

\begin{table}[h!]
    \centering
    \begin{tabular}{c cccc}
        \toprule
        & \multicolumn{4}{c}{Hyperparameters} \\
        & $\ROMrhsliquid$/$\ROMobs$ neurons & $\ROMobs$ layers & $N_f$ & $N_s$ \\
        \midrule
        tuning & $\{200, 250, \; ... \;, 400\}$ & $\{5, 10, 15\}$ & $\{25, 50, \; ... \;, 200\}$ & $\{ 50, 100, 150 \}$ \\
        final    & 300 & 5 & 100 & 100 \\
        \bottomrule
    \end{tabular}
    \caption{LFLDNets hyperparameter tuning for the CFD test case. Hyperparameter ranges (top) and optimized values (bottom). The final model sizes are $\ROMrhsliquid$ (633K parameters), $\ROMobs$ (542K parameters) and $\fourier$ (300 parameters).}
    \label{tab:HPO_CFD}  
\end{table}

\begin{figure}
     \centering
     \begin{subfigure}[b]{0.46\textwidth}
         \centering
         \includegraphics[width=\textwidth]{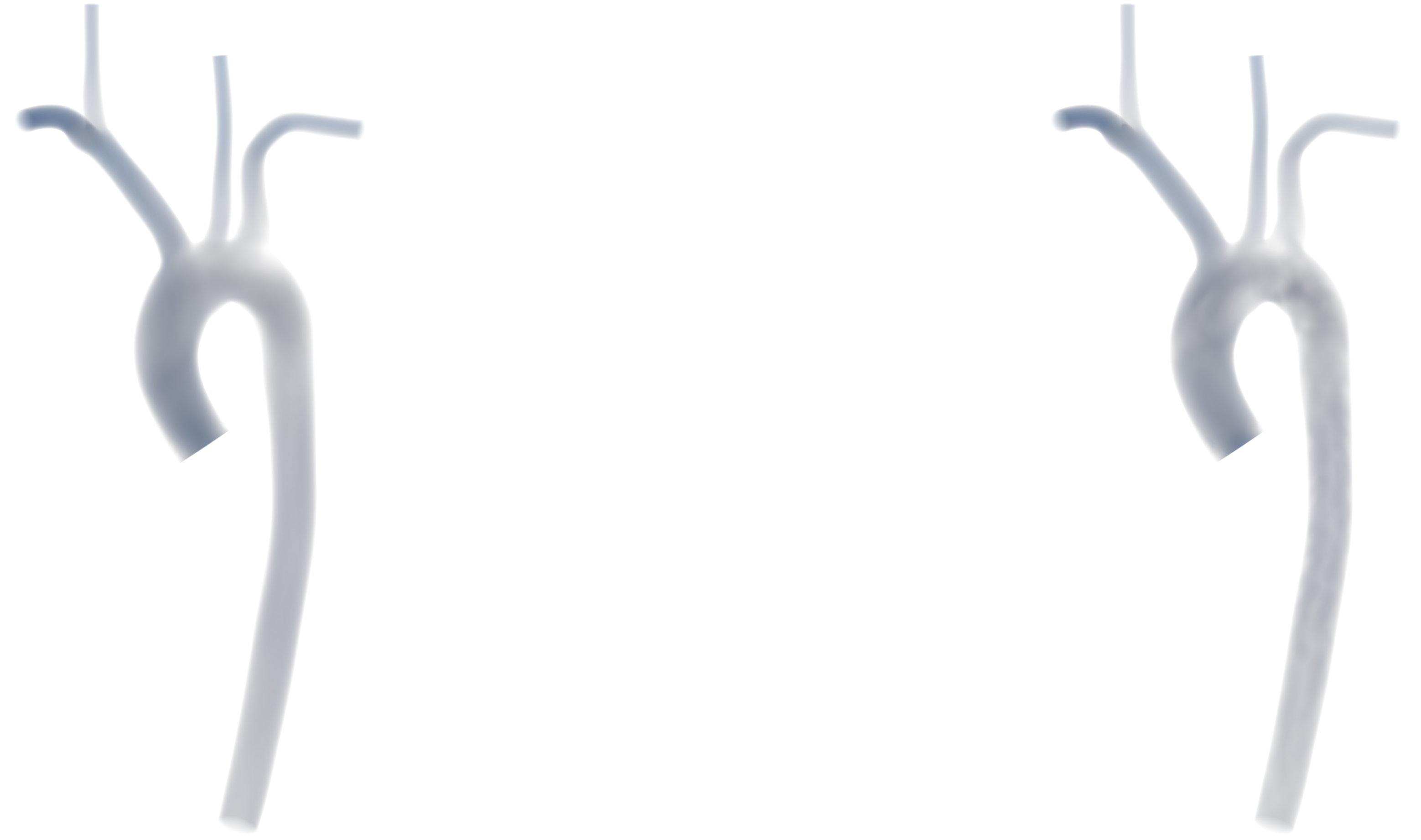}
         \caption{$t = 21$ ms}
         \label{fig:21}
     \end{subfigure}
     \hfill
     \begin{subfigure}[b]{0.46\textwidth}
         \centering
         \includegraphics[width=\textwidth]{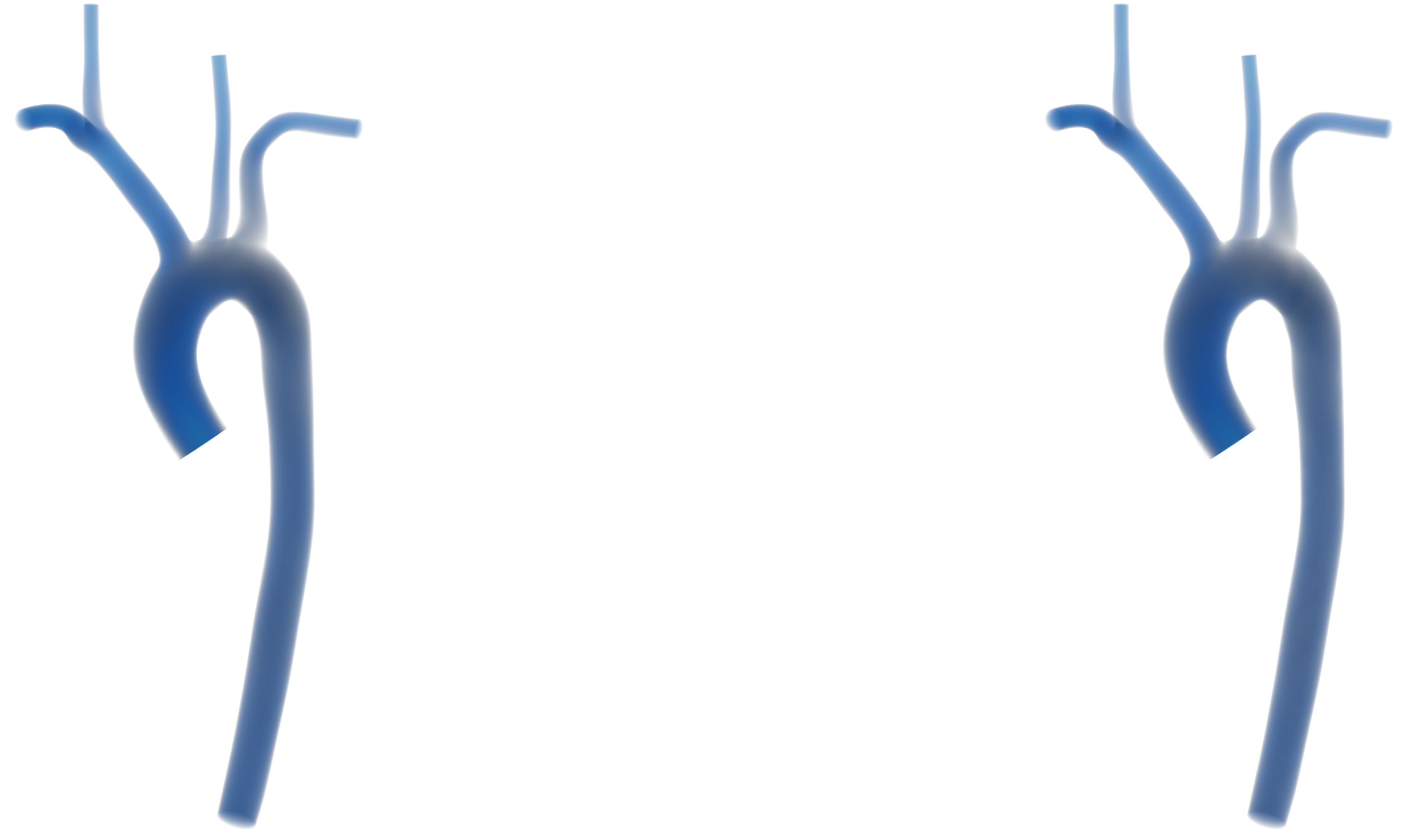}
         \caption{$t = 59.5$ ms}
         \label{fig:59.5}
     \end{subfigure}
     \hfill
     \begin{subfigure}[b]{0.46\textwidth}
         \centering
         \includegraphics[width=\textwidth]{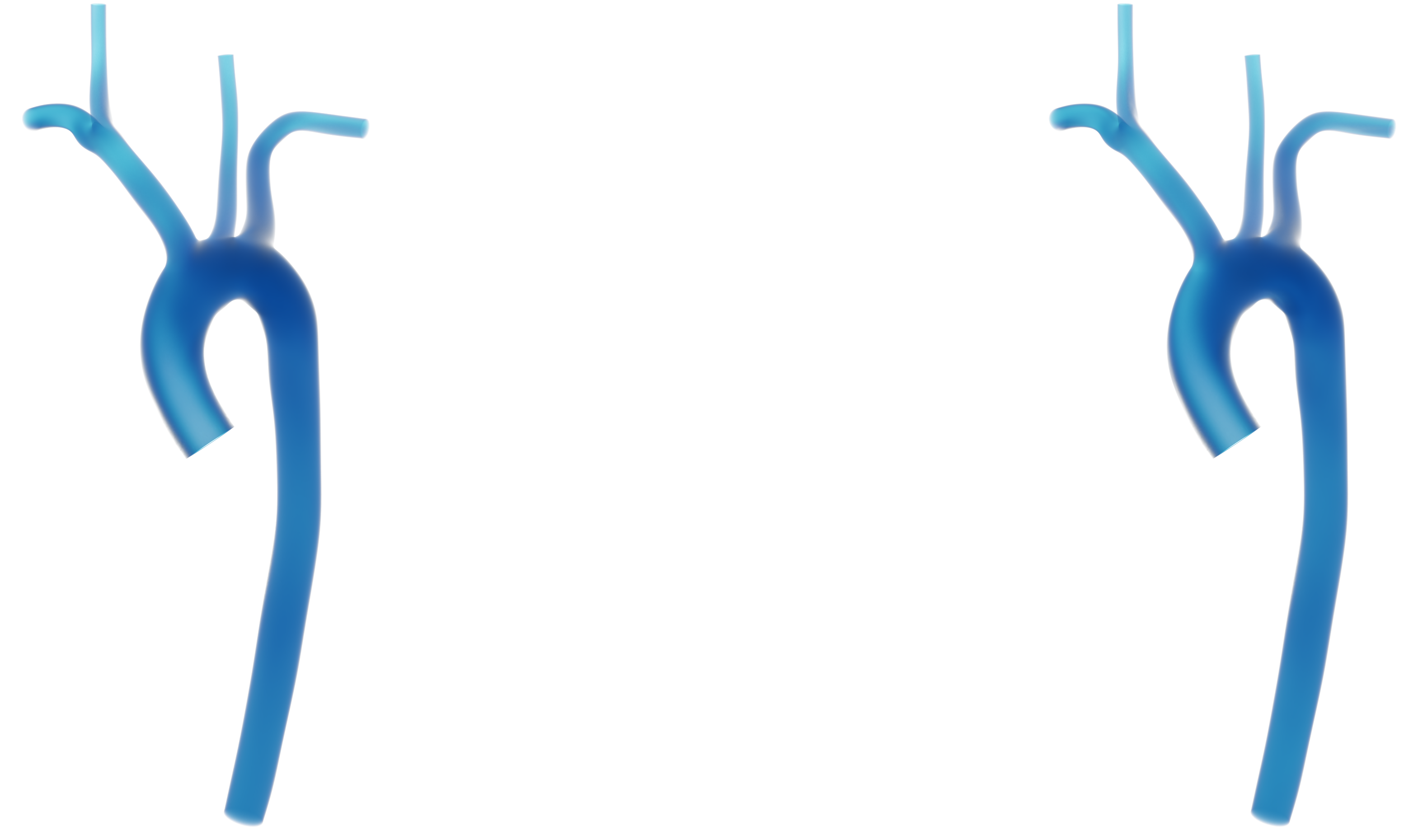}
         \caption{$t = 112$ ms}
         \label{fig:112}
     \end{subfigure}
     \hfill
     \begin{subfigure}[b]{0.46\textwidth}
         \centering
         \includegraphics[width=\textwidth]{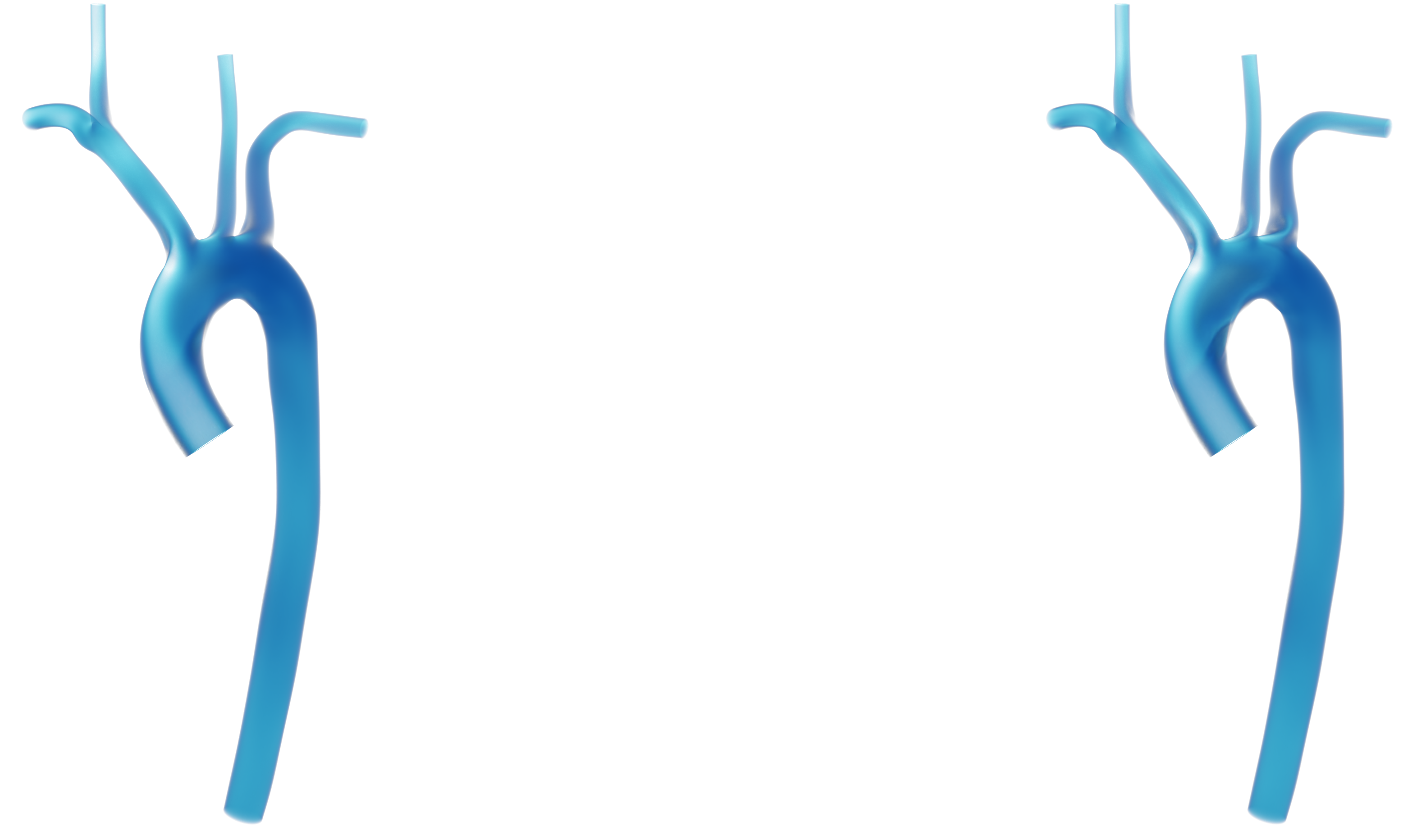}
         \caption{$t = 140$ ms}
         \label{fig:140}
     \end{subfigure}
     \hfill
     \begin{subfigure}[b]{0.5\textwidth}
         \centering
         \includegraphics[width=\textwidth]{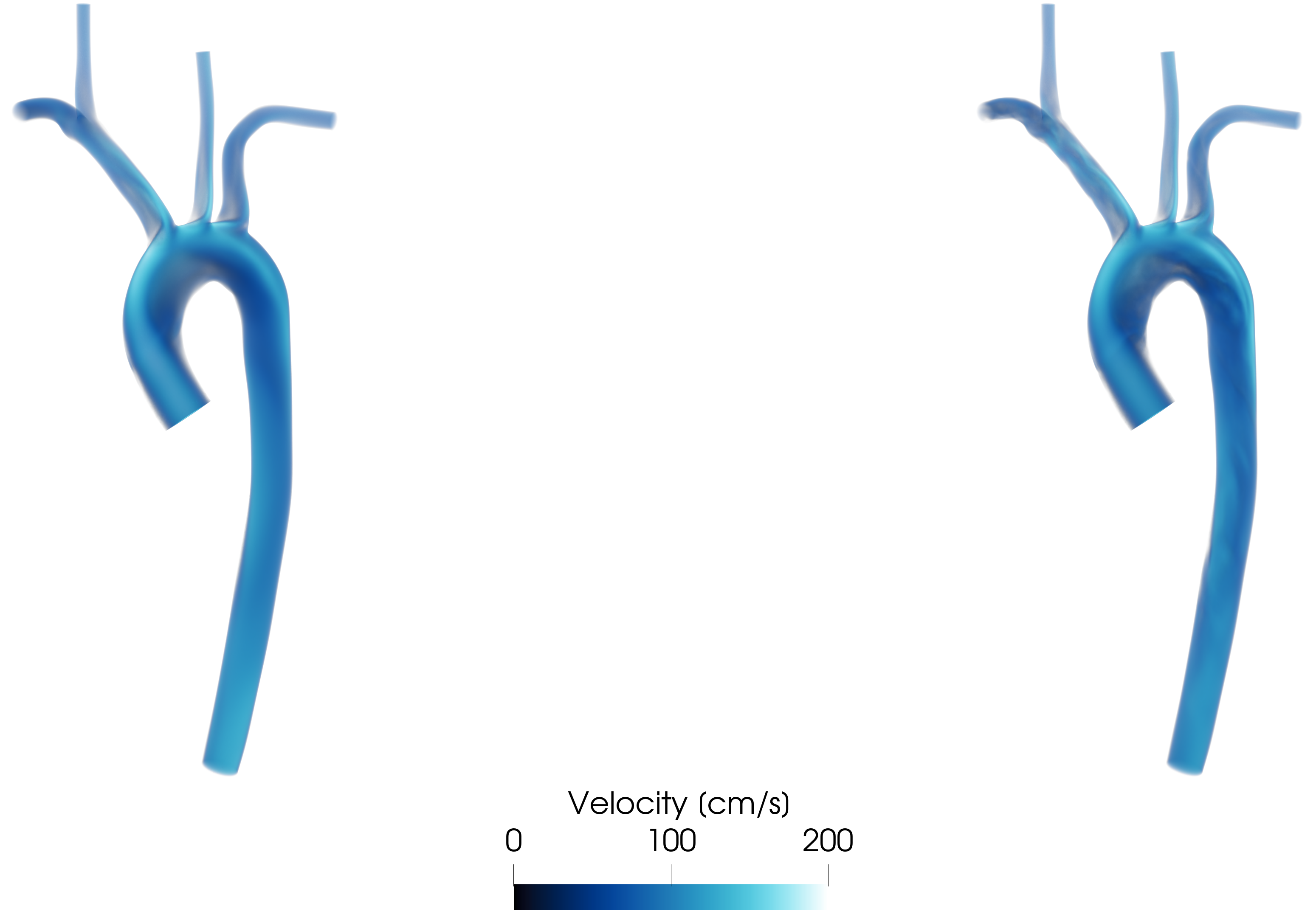}
         \caption{$t = 262.5$ ms}
         \label{fig:262.5}
     \end{subfigure}
     \caption{CFD test case. Comparison of the velocity magnitude $||\boldsymbol{u}(\spacevar, \timevar)||$ during systole for a random sample in the validation set. LFLDNet prediction (left), ground truth from CFD simulation (right).}
     \label{fig:CFD_velocity}
\end{figure}

\begin{figure}
     \centering
     \begin{subfigure}[b]{0.49\textwidth}
         \centering
         \includegraphics[width=\textwidth]{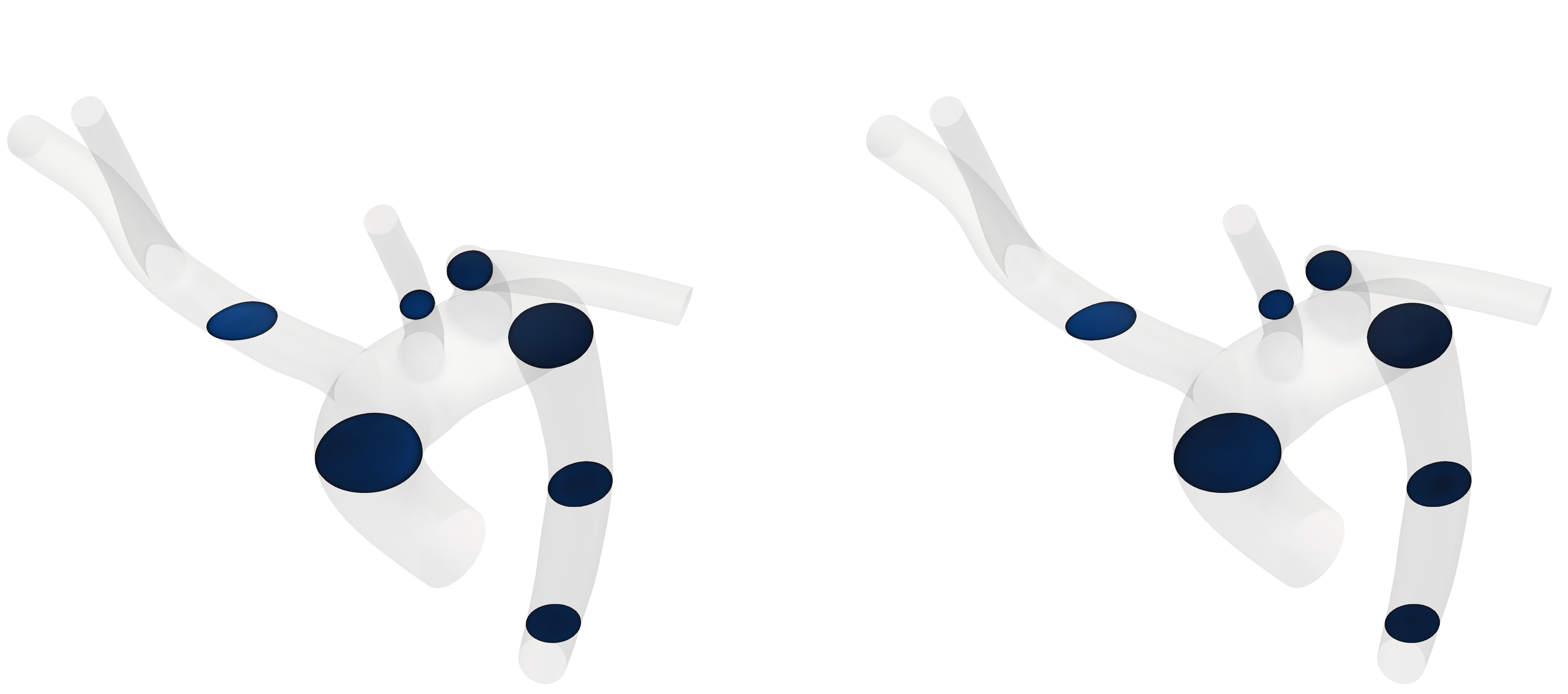}
         \caption{$t = 42$ ms}
         \label{fig:42}
     \end{subfigure}
     \hfill
     \begin{subfigure}[b]{0.49\textwidth}
         \centering
         \includegraphics[width=\textwidth]{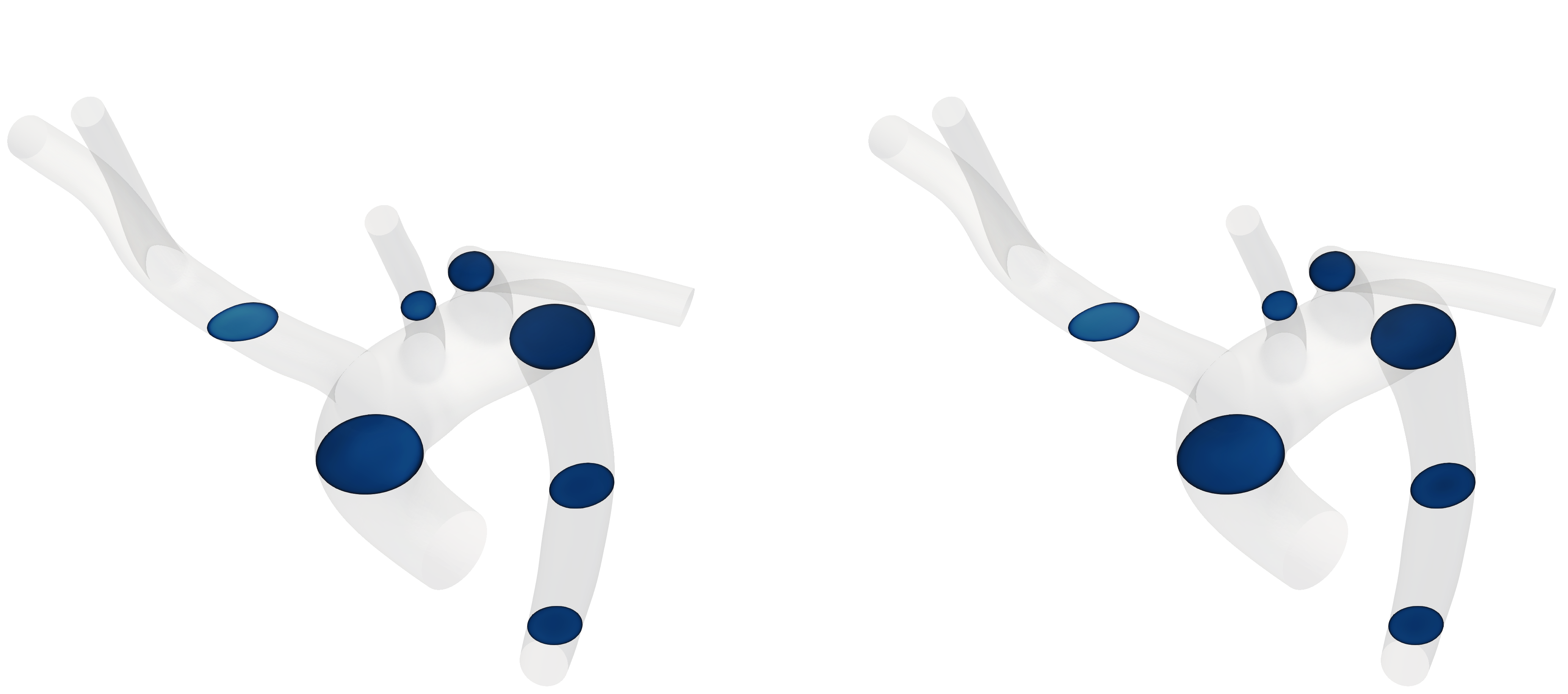}
         \caption{$t = 70$ ms}
         \label{fig:70}
     \end{subfigure}
     \hfill
     \begin{subfigure}[b]{0.49\textwidth}
         \centering
         \includegraphics[width=\textwidth]{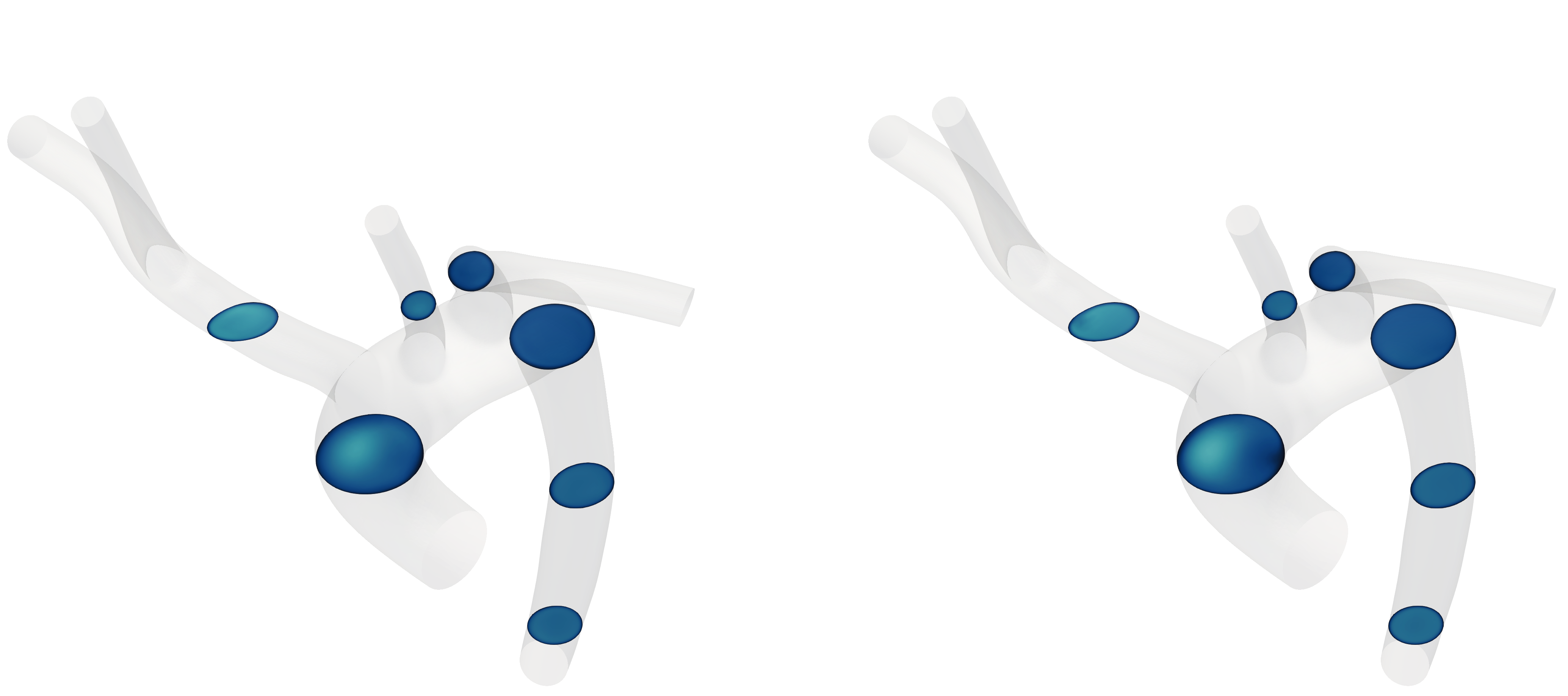}
         \caption{$t = 105$ ms}
         \label{fig:105}
     \end{subfigure}
     \hfill
     \begin{subfigure}[b]{0.49\textwidth}
         \centering
         \includegraphics[width=\textwidth]{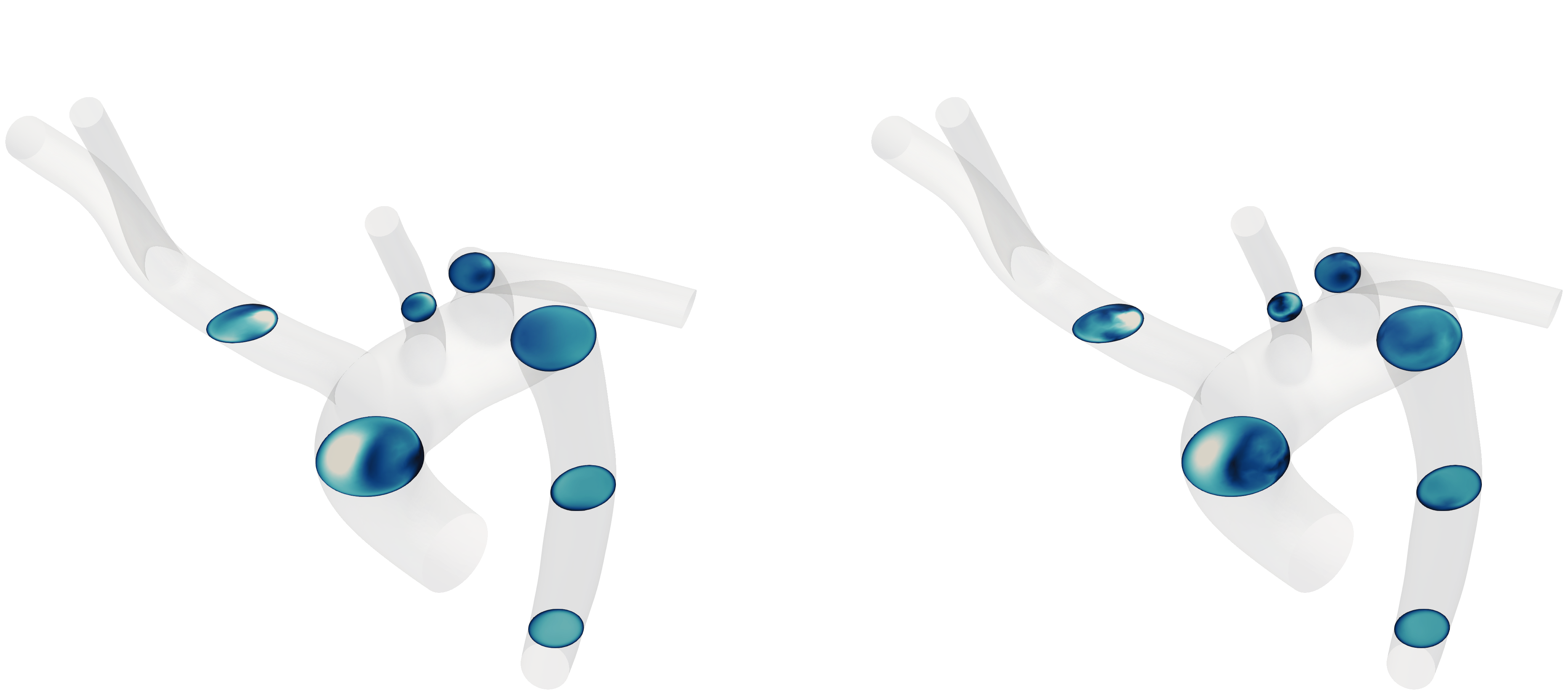}
         \caption{$t = 193$ ms}
         \label{fig:193}
     \end{subfigure}
     \hfill
     \begin{subfigure}[b]{0.49\textwidth}
         \centering
         \includegraphics[width=\textwidth]{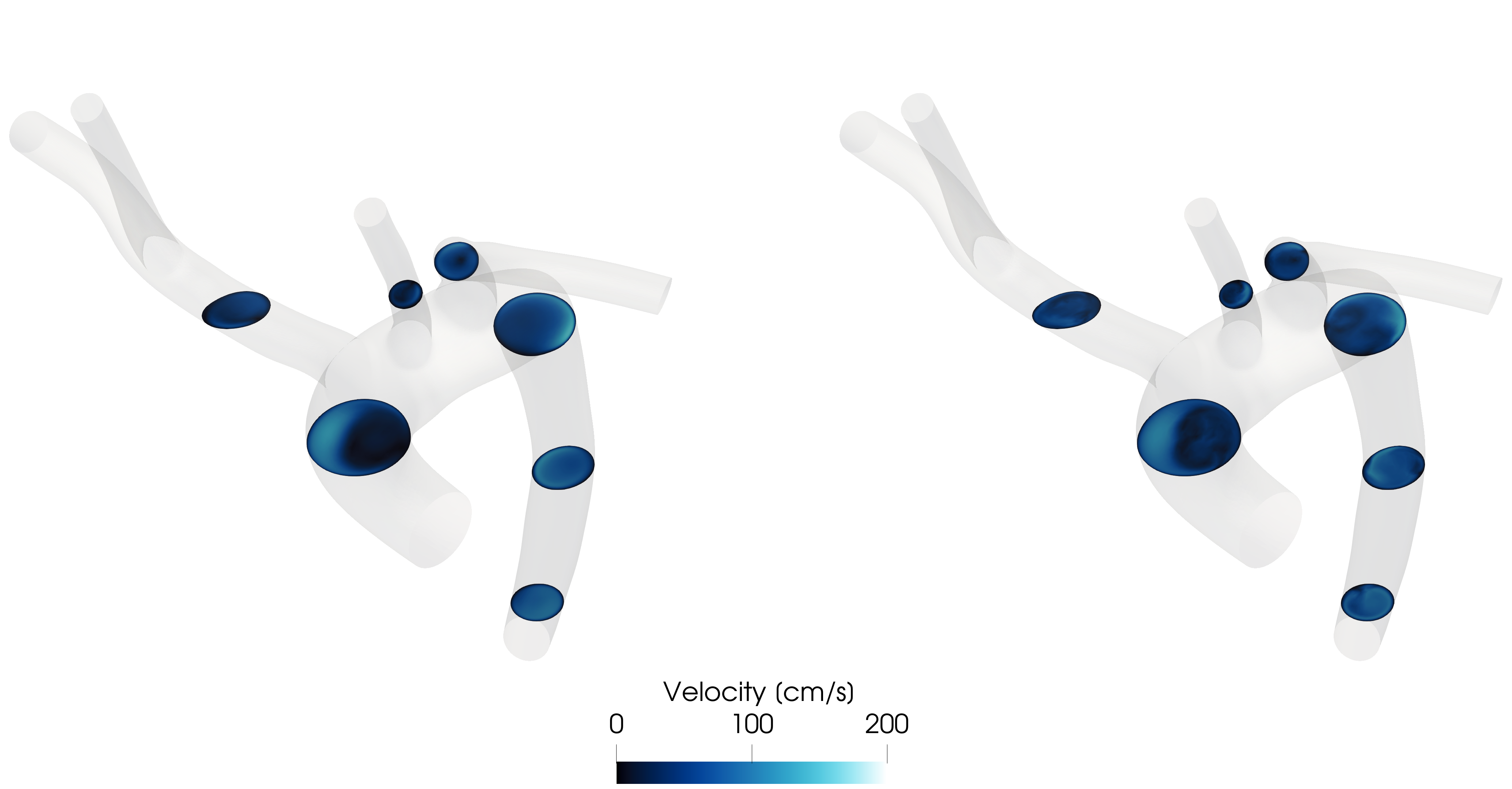}
         \caption{$t = 297.5$ ms}
         \label{fig:297.5}
     \end{subfigure}
     \caption{CFD test case. Comparison of the velocity magnitude $||\boldsymbol{u}(\spacevar, \timevar)||$ over the cardiac cycle along some slices for a random sample in the validation set. LFLDNet prediction (left), ground truth from CFD simulation (right).}
     \label{fig:CFD_slices}
\end{figure}

\begin{figure}[t!]
    \centering
    \includegraphics[width=0.8\textwidth]{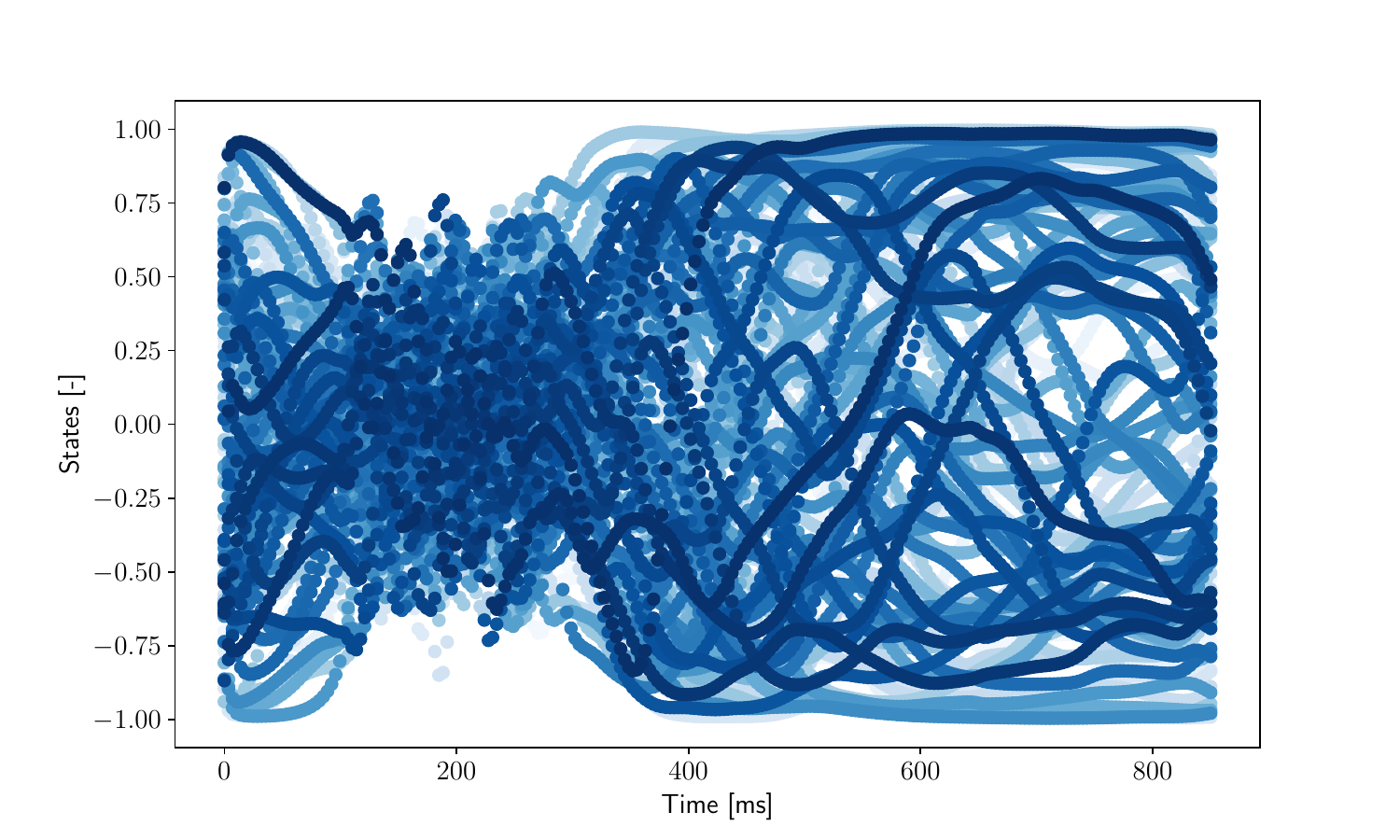}
    \caption{CFD test case. Evolution of the global state vector $\stateROM(\timevar)$ over the cardiac cycle for a specific choice of the input signals $\inputSign(\timevar)$ coming from a random sample in the validation set. Different colors identify different components of $\stateROM(\timevar)$.}
    \label{fig:CFD_states}
\end{figure}

In Table~\ref{tab:HPO_CFD} we show the tuning ranges and final values of the LFLDNets hyperparameters after running the TPE algorithm.
Although we consider a challenging CFD test case, we still obtain an optimal LFLDNet that is quite small, on the order of a million of tunable parameters.
It should be noted that by considering only the final heartbeat, which is closer to the limit cycle, the LFLDNet spans different initial conditions for each numerical simulation due to different choices of boundary conditions from Table~\ref{tab:CFD:parameterspace}, which further increases the overall training complexity.

According to the specific architecture, hyperparameter tuning for LFLDNets over 500 epochs requires from 30 minutes to 9 hours.
The optimal configuration achieves convergence at epoch 5.599 after 6 hours of computation, with a normalized MSE of 3.32e-4 over the training set (25 simulations) and 5.25e-4 for the validation set (7 simulations).
Running inference on a spatio-temporal, numerical simulation in eager (rather than graph) execution requires about 45 minutes.
We note that these computational times can be significantly reduced by training or querying LFLDNets with a coarser spatio-temporal resolution, as here we consider 240 snapshots over 560,868 DOFs compared to the 60 time points over 240,555 DOFs of the electrophysiology test case.
A high-fidelity simulation takes about 3 to 4 hours on 128 cores \cite{Pegolotti2024}.
In Figures~\ref{fig:CFD_velocity} and~\ref{fig:CFD_slices} we depict the spatio-temporal evolution of the velocity magnitude over the cardiac cycle for a random sample of the validation set.
We remark that, during both training and inference, we apply a mask over $\Gamma^\mathrm{wall}$ to strongly impose homogeneous Dirichlet boundary conditions within the LFLDNet.
We see that the LFLDNet output closely match the high-fidelity numerical simulation, although small eddies are not well captured throughout the cardiac cycle and the approximation error in the arteries is higher than that observed along the ascending and descending aorta.
This can also be motivated by the small training set, consisting of only 25 simulations while covering a large parameter space of plausible initial and boundary conditions, which is nevertheless sufficient to ensure small generalization errors and to correctly capture large to medium-sized eddies thanks to all the mathematical properties of LFLDNets.
In fact, a larger training set, combined with bigger dynamics and reconstruction networks, a higher number of states (and Fourier features) would potentially allow for better performance \cite{Regazzoni2024}.
Furthermore, we stress that the uniform time step chosen for training and inference, i.e. $\Delta t=0.0035$ s, is higher than that allowed by the Courant-Friedrichs-Lewy (CFL) condition.
As mentioned in the Methods section, all the mathematical operations involving LFLDNets run on 1 Nvidia A40 GPU.

In Figure~\ref{fig:CFD_states} we depict the trajectories of the global state vector $\stateROM(\timevar)$ coming from the optimal LFLDNet for a random sample of the validation set.
Again, we notice that these 100 states present a smooth temporal dynamics that uniformly span the bounded [-1, 1] non-dimensional range.
However, significant variability within $\stateROM(\timevar)$ occurs in the time interval $[100, 300]$ ms, i.e., during the ejection part of the systolic phase, where jets, complex flow patterns, and a high velocity field are experienced.


	\section{Discussion}
\label{sec:discussion}

We proposed LFLDNets, a novel neural operator that extends the capabilities of LDNets \cite{Regazzoni2024} in key ways, to create reduced-order surrogate models capable of capturing the spatio-temporal dynamics of parameterized time-dependent PDEs.
We considered two challenging 3D test cases in computational cardiology to show their advantages and features, and assess performance.

LFLDNets uses a CfC, NCP dynamics network to define the time evolution of a global state vector characterizing the dynamical system \cite{Hasani2022CfC, Lechner2020NCP}.
Unlike neural ODEs, LNNs exploit a closed-form solution of the ODE system and let the state vector evolve without a specific temporal discretization scheme.
Furthermore, these sparse neural networks are at least an order of magnitude faster and more expressive than neural ODEs, while requiring fewer tunable parameters than other fully-connected architectures, thus providing improved accuracy and computational efficiency \cite{Chen2019, Hasani2022CfC}.

Similar to LDNets, the second component of LFLDNets is made by a FCNN, called reconstruction network, to recover a generic space-time field of interest \cite{Regazzoni2024}.
However, in LFLDNets, the spatial coordinates provided as input to the reconstruction network are fed into a trainable Fourier encoding so as to prioritize the learning of high-frequency functions over smooth features during training.
This allows the model to represent a wider range of frequency content than prior methods with a slight increase in computational cost. 

CfC, NCP dynamics networks provided superior convergence during training along with smooth temporal dynamics that is also broader, better distributed, and more representative of the physical system compared to the neural ODE counterpart.
In fact, as also observed in \cite{Lim2024,Salvador2024BLNM}, enforcing sparsity and breaking symmetries in the inherent structure of a neural network played a crucial role in the optimization process to determine a smoother landscape of local minima and a more monotonic decay of the loss function.

The generalization error decreased as the dimensionality of the Fourier embedding increases.
As shown in \cite{Regazzoni2024}, a similar trend can also be observed when increasing the number of states.
We also found that increasing the number of Fourier frequencies allowed us to obtain a faster decay of the loss function, which also start from lower values since the very first epochs, at the expense of having a larger input layer in the reconstruction network combined with a larger Fourier embedding.

In terms of activation functions, the use of LeCun hyperbolic tangent in the dynamics network and GELU in the reconstruction network resulted in a better and smoother decay of the loss function than pure hyperbolic tangent in both neural networks.
For the architectures explored in this paper, these relaxed the need for a learning rate scheduler in the Adam optimizer and corresponding hyperparameter tuning during the training phase.

Unlike LDNets, the use of the first-order stochastic Adam optimizer alone achieved low generalization errors and did not hinder convergence with respect to its combination with a second-order deterministic optimizer, such as L-BFGS \cite{Liu1989}, which would also have been particularly challenging and expensive during training for the architectures of this paper, which are on the order of a million of tunable parameters.

We also observed a negligible performance loss when switching from double-precision to single-precision training.
With respect to LDNets, this allowed us to speed up the computation and reduce the memory requirements, which is especially important for GPU training and bigger neural networks.
However, we remark that all wall times reported in this paper refer to eager mode execution.
This means that there is still room for improvement during training and inference in terms of overall computational time by switching to graph execution.

We tried to add physics-based terms, such as the imposition of a divergence-free velocity field in the CFD test case, to the data-driven loss function based on the MSE, but did not observe a significant improvement in accuracy.
This is motivated by the access to the full spatio-temporal numerical solution during the training phase.
Furthermore, the summation of a physics-based loss term, which encodes the strong form of a differential equation, involves an additional computational burden where the space and time derivatives are evaluated numerically using automatic differentiation.
On the other hand, a non-intrusive scientific ML method provides a faster tool where specific PDEs primal variables can be targeted during training, such as the velocity field but not the pressure in the Navier-Stokes equations, using a standard data-driven loss function that is agnostic to the actual application.

We also explored different architectures for the dynamics network.
Thanks to their remarkable generalization properties in many different applications, we tried to use transformers for the dynamics network \cite{Vaswani2017, Li2023Transformers}.
However, this method, in its original form, has higher spatio-temporal computational complexity during training and inference compared to LNNs, lack interpretability, which is especially important in computational medicine, and require larger training datasets to show good performance.
While this is generally not a limitation in other domains, such as language or vision, scientific computing may require running complex numerical simulations via high-performance computing, and generating arbitrarily large datasets is often not a viable option.
Nevertheless, if properly trained on large datasets, transformer-based models are better at learning very long temporal sequences than LNNs \cite{Hasani2022CfC}.
Moreover, overparameterized transformer-based neural networks show good out-of-distribution properties after a long training process that goes beyond the classical optimal point where validation loss starts to increase while training loss continues to decrease \cite{Wang2024}.

Similarly, we tested different reconstruction networks.
First, we considered implicit neural representations with periodic activation functions, but we did not observe better performance than the Fourier embedding \cite{Sitzmann2020}.
Second, we employed graph neural networks, which generally increased the computational cost during training and inference and did not provide better convergence properties compared to a FCNN.
Nonetheless, further investigation is certainly needed, as this tool can be very flexible in generalizing to different cardiovascular geometries \cite{Pegolotti2024}, which is not considered in this work.

With respect to other neural operators \cite{Rahman2024, Li2020, Li2023, Hao2023, Lu2021, Vlachas2022, Raonic2023}, we showed that LFLDNets are very lightweight and can be seamlessly trained in the physical space on complex geometries.
This is not always feasible, such as with Fourier neural operators \cite{Li2021}, which require a structured grid and can only work on unstructured meshes by introducing a structured reparameterization in a latent space \cite{Li2023GeoFNO}.
Furthermore, our method allows to span generic sets of initial and boundary conditions, as well as physics-based model parameters, forcing terms, space, time, space-time coefficients underpinning multiscale and multiphysics sets of PDEs.

    \section{Conclusions}
\label{sec:conclusions}

We introduce an extension of LDNets, namely LFLDNets, to improve and scale space-time surrogate models of arbitrary physical processes based on PDEs in terms of accuracy, computational efficiency, memory requirements and model dimension.
This is achieved by combining a neurologically-inspired, liquid dynamics network in time with a reconstruction network in space enriched by a Fourier embedding.
We challenge LFLDNets on two different test cases in pediatrics arising from cardiac electrophysiology for congenital heart disease and cardiovascular hemodynamics for a healthy aorta.
In both cases, LFLDNets demonstrate that significantly smaller dynamics and reconstruction networks reach lower generalization errors, improved temporal dynamics of the state vector, together with faster training and inference, compared to prior approaches, such as LDNets or LLDNets, i.e. LFLDNets without the Fourier encoding for space coordinates.

Future developments of this work should further improve the performance of the proposed method as well as the range of possible applications in engineering and digital twinning.
Some examples could be reinterpreting multifidelity domain decomposition approaches \cite{Howard2023, Heinlein2024} or recent advances in local neural operators \cite{LiuSchiaffini2024} within LFLDNets to further reduce the surrogate modeling error, incorporating recent discoveries in expressive and interpretable Kolmogorov-Arnold networks \cite{Liu2024} and related extensions \cite{Shukla2024} into the reconstruction network, as they also allow for continual learning, and encoding different geometries into LFLDNets.

	\section*{Acknowledgements}

This project has been funded by the Benchmark Capital Fellowship from the Vera Moulton Wall Center at Stanford, the SimCardio NSF grant 1663671 and the NIH grant R01HL173845.
We thank Dr. Luca Pegolotti for providing the dataset of numerical simulations for the CFD test case and for his comments on the manuscript.
We also thank Prof. Francesco Regazzoni for his feedback on LFLDNets.

    \newpage
    \printbibliography

@article {Krishnamoorthi2013,
    title = "Numerical quadrature and operator splitting in finite element methods for cardiac electrophysiology",
    journal = "International Journal for Numerical Methods in Biomedical Engineering",
    author = "Krishnamoorthi, S. and Sarkar, M. and Klug, W. S.",
    year = "2013",
    volume = "29",
    issue = "11",
    pages = "1243-1266"
}

@article{Bayer2012,
	title = {A novel rule-based algorithm for assigning myocardial fiber orientation to computational heart models},
	journal = {Annals of Biomedical Engineering},
	volume = {40},
	author = {Bayer, J. D. and Blake, R. C. and Plank, G. and Trayanova, N.},
	year = {2012},
	pages = {2243--2254}
}

@article{regazzoni2019modellearning,
    title = {Machine learning for fast and reliable solution of time-dependent differential equations},
    journal = {Journal of Computational Physics},
    volume = {397},
    pages = {108852},
    year = {2019},
    author = {Regazzoni, F. and Dede', L. and Quarteroni, A.}
}

@book{Quarteroni2019,
    title = {Mathematical Modelling of the Human Cardiovascular System: Data, Numerical Approximation, Clinical Applications},
    publisher = {Cambridge University Press},
    author = {Quarteroni, A. and Dede', L. and Manzoni, A. and Vergara, C.},
    year = {2019}
}

@article{TTP06,
	title = {Alternans and spiral breakup in a human ventricular tissue model},
	journal = {American Journal of Physiology. Heart and Circulatory Physiology},
	volume = {291},
	author = {ten Tusscher, K. H. and Panfilov, A. V.},
	year = {2006},
	pages = {1088--1100}
}

@article{Piersanti2021,
    title = {Modeling cardiac muscle fibers in ventricular and atrial electrophysiology simulations},
    journal = {Computer Methods in Applied Mechanics and Engineering},
    volume = {373},
    pages = {113468},
    year = {2021},
    author = {Piersanti, R. and Africa, P. C. and Fedele, M. and others},
}

@article{Regazzoni2022,
    title = {A cardiac electromechanical model coupled with a lumped-parameter model for closed-loop blood circulation},
    journal = {Journal of Computational Physics},
    volume = {457},
    pages = {111083},
    year = {2022},
    author = {F. Regazzoni and M. Salvador and P. C. Africa and M. Fedele and L. Dede' and A. Quarteroni}
}

@article{Piersanti2022,
    title = {{3D-0D} closed-loop model for the simulation of cardiac biventricular electromechanics},
    journal = {Computer Methods in Applied Mechanics and Engineering},
    volume = {391},
    pages = {114607},
    year = {2022},
    author = {R. Piersanti and F. Regazzoni and M. Salvador and others}
}

@article{Costabal2016,
    journal={Journal of biomechanics},
    year={2016},
    volume={49},
    pages={2455--2465},
    title={Generating Purkinje networks in the human heart},
    author={Sahli Costabal, F. and Hurtado, D. E. and Kuhl, E.}
}

@article{Gerach2021,
    author = {Gerach, T. and Schuler, S. and Fr{\"o}hlich, J. and others},
    title = {Electro-Mechanical Whole-Heart Digital Twins: A Fully Coupled Multi-Physics Approach},
    journal = {Mathematics},
    volume = {9},
    year = {2021},
    number = {11}
}

@article{Chen2019,
    journal={arXiv:1806.07366},
    year={2019},
    title={Neural Ordinary Differential Equations},
    author={Chen, R. T. Q. and Rubanova, Y. and Bettencourt, J. and Duvenaud, D.}
}

@article{Dupont2019,
    journal={arXiv:1904.01681},
    year={2019},
    title={Augmented Neural ODEs},
    author={Dupont, E. and Doucet, A. and Teh, Y. W.}
}

@book{collifranzone2014book,
    title={Mathematical cardiac electrophysiology},
    author={Colli Franzone, P. and Pavarino, L.F. and Scacchi, S.},
    volume={13},
    year={2014},
    publisher={Springer}
}

@article{Liu1989,
    title={On the limited memory {BFGS} method for large scale optimization},
    author={Liu, D.C. and Nocedal, J.},
    journal={Mathematical Programming},
    volume={45},
    pages={503--528},
    year={1989}
}

@article{Jung2022,
    author={Jung, A. and Gsell, M. A. F. and Augustin, C. M. and Plank, G.},
    title={An Integrated Workflow for Building Digital Twins of Cardiac Electromechanics-A Multi-Fidelity Approach for Personalising Active Mechanics},
    journal={Mathematics},
    volume={10},
    year={2022},
    number={5}
}

@article{Kingma2014,
    title = {Adam: A Method for Stochastic Optimization},
    author = {Kingma, D. P. and Ba, J.},
    journal = {arXiv:1412.6980},    
    year = {2014}
}

@article{Fedele2023,
    title = {A comprehensive and biophysically detailed computational model of the whole human heart electromechanics},
    journal = {Computer Methods in Applied Mechanics and Engineering},
    volume = {410},
    pages = {115983},
    year = {2023},
    author = {Fedele, M. and Piersanti, R. and Regazzoni, F. and Salvador, M. and Africa, P. C. and Bucelli, M. and Zingaro, A. and Dede', L. and Quarteroni, A.}
}

@inproceedings{Optuna2019,
    title = {Optuna: A Next-generation Hyperparameter Optimization Framework},
    author = {Akiba, T. and Sano, S. and Yanase, T. and Ohta, T. and Koyama, M.},
    booktitle = {Proceedings of the 25rd {ACM} {SIGKDD} International Conference on Knowledge Discovery and Data Mining},
    year = {2019}
}

@article{Bergstra2011,
  title = {Algorithms for hyper-parameter optimization},
  author = {Bergstra, J. and Bardenet, R. and Bengio, Y. and K{\'e}gl, B.},
  journal = {Advances in neural information processing systems},
  volume = {24},
  year = {2011}
}

@inproceedings{Ray2018,
    author = {Moritz, P. and Nishihara, R. and Wang, S. and Tumanov, A. and Liaw, R. and Liang, E. and Elibol, M. and Yang, Z. and Paul, W. and Jordan, M. I. and Stoica, I.},
    title = {Ray: A Distributed Framework for Emerging AI Applications},
    year = {2018},
    booktitle = {Proceedings of the 13th USENIX Conference on Operating Systems Design and Implementation},
    pages = {561--577}
}

@article{Niederer2019Nature,
    author = {Niederer, S. A. and Lumens, J. and Trayanova, N. A.},
    title = {Computational models in cardiology},
    journal = {Nature Reviews Cardiology},
    volume = {16},
    pages = {100-111},
    year = {2019}
}

@article{CorralAcero2020,
    author = {Corral-Acero, J. and Margara, F. and Marciniak, M. and Rodero, C. and et al.},
    title = "{The ‘Digital Twin’ to enable the vision of precision cardiology}",
    journal = {European Heart Journal},
    volume = {41},
    number = {48},
    pages = {4556-4564},
    year = {2020}
}

@article{Regazzoni2024,
    journal={Nature Communications},
    volume={15},
    year={2023},
    title={Learning the intrinsic dynamics of spatio-temporal processes through Latent Dynamics Networks},
    author={Regazzoni, F. and Pagani, S. and Salvador, M. and Dede', L. and Quarteroni, A.}
}

@article{Salvador2024EMROM4CH,
    journal={npj Digital Medicine},
    volume={7},
    year={2024},
    title={Whole-heart electromechanical simulations using Latent Neural Ordinary Differential Equations},
    author={Salvador, M. and Strocchi, M. and Regazzoni, F. and Augustin, C. M. and Dede', L. and Niederer, S. A. and Quarteroni, A.}
}

@article{Pegolotti2024,
    journal={Computers in Biology and Medicine},
    year={2024},
    title={Learning reduced-order models for cardiovascular simulations with graph neural networks},
    author={Pegolotti, L. and Pfaller, M. R. and Rubio, N. L. and Ding, K. and Brufau, R. B. and Darve, E. and Marsden, A. L.}
}

@article{Gillette2021,
    title = "{A Framework for the generation of digital twins of cardiac electrophysiology from clinical 12-leads ECGs}",
    journal = {Medical Image Analysis},
    volume = {71},
    pages = {102080},
    year = {2021},
    author = {Gillette, K. and Gsell, M.A.F. and Prassl, A.J. and Karabelas, E. and Reiter, U. and Reiter, G. and Grandits, T. and Payer, C. and Štern, D. and Urschler, M. and Bayer, J.D. and Augustin, C.M. and Neic, A. and Pock, T. and Vigmond, E.J. and Plank, G.}
}

@article{Zhu2022,
  year = {2022},
  publisher = {The Open Journal},
  volume = {7},
  number = {78},
  pages = {4118},
  author = {Zhu, C. and Vedula, V. and Parker, D. and Wilson, N. and Shadden, S. and Marsden, A.},
  title = {svFSI: A Multiphysics Package for Integrated Cardiac Modeling}, journal = {Journal of Open Source Software}
}

@article{Tikenogullari2023,
  title = {Effects of cardiac growth on electrical dyssynchrony in the single ventricle patient},
  journal = {Computer Methods in Biomechanics and Biomedical Engineering},
  volume = {0},
  number = {0},
  pages = {1-17},
  year  = {2023},
  publisher = {Taylor & Francis},
  author = {Tikenogullari, O. Z. and Peirlinck, M. and Chubb, H. and Dubin, A. M. and Kuhl, E. and Marsden, A. L.}
}

@article{Lu2021,
  author={Lu, L. and Jin, P. and Pang, G. and Zhang, Z. and Karniadakis, G. E.},
  journal={Nature Machine Intelligence}, 
  title={Learning nonlinear operators via DeepONet based on the universal approximation theorem of operators}, 
  year={2021},
  volume={3},
  pages={218-229}
}

@article{Vlachas2022,
  title={Multiscale simulations of complex systems by learning their effective dynamics},
  author={Vlachas, P. R. and Arampatzis, G. and Uhler, C. and Koumoutsakos, P.},
  journal={Nature Machine Intelligence},
  volume={4},
  number={4},
  pages={359--366},
  year={2022}
}

@conference{Raonic2023,
  author={Raonic, B. and Molinaro, R. and Rohner, T. and Mishra, S. de Bezenac, E.},
  booktitle={ICLR Workshop on Physics for Machine Learning}, 
  title={Convolutional Neural Operators}, 
  year={2023}
}

@article{Salvador2024BLNM,
  title={Branched Latent Neural Maps},
  author={Salvador, M. and Marsden, A. L.},
  journal={Computer Methods in Applied Mechanics and Engineering},
  volume={418},
  pages={116499},
  year={2024}
}

@article{Updegrove2017,
    title = {SimVascular: An Open Source Pipeline for Cardiovascular Simulation},
    journal = {Annals of Biomedical Engineering},
    volume = {45},
    pages = {525–541},
    year = {2017},
    author = {Updegrove, A. and Wilson, N. M. and Merkow, J. and Lan, H. and Marsden, A. L. and Shadden, S. C.}
}

@article {Kong2023,
    author = {Kong, F. and Stocker, S. and Choi, P. S. and Ma, M. and Ennis, D. B. and Marsden, A.},
    title = {SDF4CHD: Generative Modeling of Cardiac Anatomies with Congenital Heart Defects},
    journal = {arXiv:2311.00332},
    year = {2023}
}

@article{Viola2023,
    title = {GPU accelerated digital twins of the human heart open new routes for cardiovascular research},
    journal = {Scientific Reports},
    volume = {13},
    year = {2023},
    author = {Viola, F. and Del Corso, G. and De Paulis, R. and Verzicco, R.}
}

@article{Howard2023,
    title = {Multifidelity Deep Operator Networks For Data-Driven and Physics-Informed Problems},
    journal = {Journal of Computational Physics},
    pages = {112462},
    year = {2023},
    author = {Howard, A. A. and Perego, M. and Karniadakis, G. E. and Stinis, P.}
}

@article{Li2021,
    journal={International Conference on Learning Representations},
    year={2021},
    title={Fourier Neural Operator for Parametric Partial Differential Equations},
    author={Li, Z. and Kovachki, N. and Azizzadenesheli, K. and Liu, B. and Bhattacharya, K. and Stuart, A. and Anandkumar, A.}
}

@article{Rahman2024,
    journal={arXiv:2403.12553},
    year={2024},
    title={Pretraining Codomain Attention Neural Operators for Solving Multiphysics PDEs},
    author={Rahman, M. A. and George, R. J. and Elleithy, M. and Leibovici, D. and Li, Z. and Bonev, B. and White, C. and Berner, J. and Yeh, R. A. and Kossaifi, J. and Azizzadenesheli, K. and Anandkumar, A.}
}

@article{Li2020,
    journal={Advances in Neural Information Processing Systems},
    year={2020},
    title={Multipole Graph Neural Operator for Parametric Partial Differential Equations},
    author={Li, Z. and Kovachki, N. and Azizzadenesheli, K. and Liu, B. and Stuart, A. and Bhattacharya, K. and Anandkumar, A.}
}

@article{Sitzmann2020,
    journal={arXiv:2006.09661},
    year={2020},
    title={Implicit neural representations with periodic activation functions},
    author={Sitzmann, V. and Martel, J. N. P. and Bergman, A. W. and Lindell, D. B. and Wetzstein, G.}
}

@article{Tancik2020,
    journal={arXiv:2006.10739},
    year={2020},
    title={Fourier features let networks learn high frequency functions in low dimensional domains},
    author={Tancik, M. and Srinivasan, P. P. and Mildenhall, B. and Fridovich-Keil, S. and Raghavan, N. and Singhal, U. and Ramamoorthi, R. and Barron, J. T. and Ng, R.}
}

@article{Hennigh2020,
    journal={International Conference on Computational Science},
    year={2020},
    title={NVIDIA SimNet: An AI-Accelerated Multi-Physics Simulation Framework},
    author={Hennigh, O. and Narasimhan, S. and Nabian, M. A. and Subramaniam, A. and Tangsali, K. and Fang, Z. and Rietmann, M. and Byeon, W. and Choudhry, S.}
}

@article{Li2023,
    journal={Neural Information Processing Systems},
    year={2023},
    title={Geometry-Informed Neural Operator for Large-Scale 3D PDEs},
    author={Li, Z. and Kovachki, N. and Choy, C. and Li, B. and Kossaifi, J. and Otta, S. P. and Nabian, M. A. and Stadler, M. and Hundt, C. and Azizzadenesheli, K. and Anandkumar, A.}
}

@article{Hao2023,
    journal={Proceedings of the $40^{th}$ International Conference on Machine Learning},
    year={2023},
    title={GNOT: A General Neural Operator Transformer for Operator Learning},
    author={Hao, Z. and Wang, Z. and Su, H. and Ying, C. and Dong, Y. and Liu, S. and Cheng, Z. and Song, J. and Zhu, J.}
}

@article{Li2023Transformers,
    journal={Transactions on Machine Learning Research},
    year={2023},
    title={Transformer for Partial Differential Equations' Operator Learning},
    author={Li, Z. and Meidani, K. and Farimani, A. B.}
}

@article{Azizzadenesheli2024,
    journal={Nature Reviews Physics},
    year={2024},
    title={Neural operators for accelerating scientific simulations and design},
    author={Azizzadenesheli, K. and Kovachki, N. and Li, Z. and Liu-Schiaffini, M. and Kossaifi, J. and Anandkumar, A.}
}

@article{Laubenbacher2024,
    journal={Nature Computational Science},
    year={2024},
    title={Digital twins in medicine},
    volume={4},
    pages={184--191},
    author={Laubenbacher, R. and Mehrad, B. and Shmulevich, I. and Trayanova, N.}
}

@article{Willcox2024,
    journal={Nature Computational Science},
    year={2024},
    volume={4},
    pages={147--149},
    title={The role of computational science in digital twins},
    author={Willcox, K. and Segundo, B.}
}

@article{Salvador2024DTCHD,
    journal={Journal of the Royal Society Interface},
    year={2024},
    title={Digital twinning of cardiac electrophysiology for congenital heart disease},
    author={Salvador, M. and Kong, F. and Peirlinck, M. and Parker, D. W. and Chubb, H. and Dubin, A. M. and Marsden, A. L.}
}

@article{Ferrari2024,
    journal={Nature Computational Science},
    year={2024},
    volume={4},
    pages={178--183},
    title={Digital twins in mechanical and aerospace engineering},
    author={Ferrari, A. and Willcox, K.}
}

@article{Karniadakis2021,
    journal={Nature Reviews Physics},
    year={2021},
    volume={3}, 
    pages={422-–440},
    title={Physics-informed machine learning},
    author={Karniadakis, G. E. and Kevrekidis, I. G. and Lu, L. and Perdikaris, P. and Wang, S. and Yang, L.}
}

@article{Lechner2020NCP,
  title={Neural circuit policies enabling auditable autonomy},
  author={Lechner, M. and Hasani, R. and Amini, A. and Henzinger, T. A. and Rus, D. and Grosu, R.},
  journal={Nature Machine Intelligence},
  volume={2},
  number={10},
  pages={642--652},
  year={2020}
}

@article{Li2023GeoFNO,
    journal={Journal of Machine Learning Research},
    year={2023},
    volume={24},
    pages={1--26},
    title={Fourier Neural Operator with Learned Deformations for PDEs on General Geometries},
    author={Li, Z. and Huang, D. Z. and Liu, B. and Anandkumar, A.}
}

@article{Hasani2021LNN,
    journal={Proceedings of the AAAI Conference on Artificial Intelligence},
    year={2021},
    volume={35},
    pages={7657–-7666},
    title={Liquid time-constant networks},
    author={Hasani, R. and Lechner, M. and Amini, A. and Rus, D. and Grosu, R.}
}

@article{Vorbach2021LNN,
    journal={Proceedings of Advances in Neural Information Processing Systems},
    year={2021},
    pages={12425-–12440},
    title={Causal navigation by continuous-time neural networks},
    author={Vorbach, C. and Hasani, R. and Amini, A. and Lechner, M. and Rus, D.}
}

@article{Hasani2022CfC,
    title={Closed-form continuous-time neural networks},
    journal={Nature Machine Intelligence},
    author={Hasani, R. and Lechner, M. and Amini, A. and Liebenwein, L. and Ray, A. and Tschaikowski, M. and Teschl, G. and Rus, D.},
    year={2022},
    volume={4},
    pages={992–-1003}
}

@article{Liu2024,
    journal={arXiv:2404.19756},
    year={2024},
    title={KAN: Kolmogorov-Arnold Networks},
    author={Liu, Z. and Wang, Y. and Vaidya, S. and Ruehle, F. and Halverson, J. and Soljačić, M. and Hou, T. Y. and Tegmark, M.}
}

@Inbook{Goswami2023,
    author={Goswami, S. and Bora, A. and Yu, Y. and Karniadakis, G. E.},
    title={Physics-Informed Deep Neural Operator Networks},
    bookTitle={Machine Learning in Modeling and Simulation: Methods and Applications},
    year={2023},
    pages={219--254}
}

@article{He2024,
    title={Sequential Deep Operator Networks (S-DeepONet) for predicting full-field solutions under time-dependent loads},
    journal={Engineering Applications of Artificial Intelligence},
    volume={127},
    pages={107258},
    year={2024},
    author={He, J. and Kushwaha, S. and Park, J. and Koric, S. and Abueidda, D. and Jasiuk, I.}
}

@article{Whiting2001,
    author = {Whiting, C. H. and Jansen, K. E.},
    title = {A stabilized finite element method for the incompressible Navier–Stokes equations using a hierarchical basis},
    journal = {International Journal for Numerical Methods in Fluids},
    volume = {35},
    number = {1},
    pages = {93--116},
    year = {2001}
}

@article{Liu2021,
    author = {Liu, J. and Lan, I. S. and Tikenogullari, O. Z. and Marsden, A. L.},
    title = {A note on the accuracy of the generalized-$\alpha$ scheme for the incompressible Navier-Stokes equations},
    journal = {International Journal for Numerical Methods in Engineering},
    volume = {122},
    number = {2},
    pages = {638-651},
    year = {2021}
}

@article{Wilson2013,
    author = {Wilson, N. M. and Ortiz, A. K. and Johnson, A. B.},
    title = {The Vascular Model Repository: A Public Resource of Medical Imaging Data and Blood Flow Simulation Results},
    journal = {Journal of Medical Devices},
    volume = {7},
    number = {4},
    pages = {0409231},
    year = {2013}
}

@article{Pfaller2021,
    journal={Annals of Biomedical Engineering},
    year={2021},
    volume={49},
    pages={3574--3592},
    title={On the Periodicity of Cardiovascular Fluid Dynamics Simulations},
    author={Pfaller, M. R. and Pham, J. and Wilson, N. M. and Parker, D. W. and Marsden, A. L.}
}

@article{Taylor2010,
    author={Vignon-Clementel, I. E. and Figueroa, C. A. and Jansen, K. E. and Taylor, C. A.},
    title={Outflow boundary conditions for 3D simulations of non-periodic blood flow and pressure fields in deformable arteries},
    journal={Computer Methods in Biomechanics and Biomedical Engineering},
    volume={13},
    number={5},
    pages={625--640},
    year={2010}
}

@article{Meyer2008,
    journal={Toxicological Sciences},
    year={2008},
    volume={106},
    pages={5--208},
    title={Caenorhabditis elegans: An Emerging Model in Biomedical and Environmental Toxicology},
    author={Leung, M. C. K. and Williams, P. L. and Benedetto, A. and Au, C. and Helmcke, K. J. and Aschner, M. and Meyer, J. N.}
}

@article{Hendrycks2016,
  title = {Gaussian Error Linear Units (GELUs)},
  author = {Hendrycks, D. and Gimpel, K.},
  journal = {arXiv:1606.08415},
  year = {2016}
}

@inproceedings{Vaswani2017,
  author = {Vaswani, A. and Shazeer, N. and Parmar, N. and Uszkoreit, J. and Jones, L. and Gomez, A. N. and Kaiser, L. and Polosukhin, I.},
  booktitle = {Advances in Neural Information Processing Systems},
  title = {Attention is All You Need},
  volume = {30},
  year = {2017}
}

@article{Hodgkin1952,
  author = {Hodgkin, A. L. and Huxley, A. F.},
  title = {A quantitative description of membrane current and its application to conduction and excitation in nerve},
  journal = {The Journal of Physiology},
  volume = {117},
  number = {4},
  pages = {500-544},
  year = {1952}
}

@article{Shukla2024,
  title = {A comprehensive and FAIR comparison between MLP and KAN representations for differential equations and operator networks},
  author = {Shukla, K. and Toscano, J. D. and Wang, Z. and Zou, Z. and Karniadakis, G. E.},
  journal = {arXiv:2406.02917},
  year = {2024}
}

@article{Wang2024,
    journal = {arXiv:2405.15071},
    year = {2024},
    title = {Grokked Transformers are Implicit Reasoners: A Mechanistic Journey to the Edge of Generalization},
    author = {Wang, B. and Yue, X. and Su, Y. and Sun, H.}
}

@article{Heinlein2024,
    journal = {arXiv:2401.07888},
    year = {2024},
    title = {Multifidelity domain decomposition-based physics-informed neural networks and operators for time-dependent problems},
    author = {Heinlein, A. and Howard, A. A. and Beecroft, D. and Stinis, P.}
}

@article{Hochreiter1997,
    author = {Hochreiter, S. and Schmidhuber, J.},
    title = "{Long Short-Term Memory}",
    journal = {Neural Computation},
    volume = {9},
    number = {8},
    pages = {1735-1780},
    year = {1997}
}

@article{Lim2024,
    journal = {arXiv:2405.20231},
    year = {2024},
    title = {The Empirical Impact of Neural Parameter Symmetries, or Lack Thereof},
    author = {Lim, D. and Putterman, M. and Walters, R. and Maron, H. and Jegelka, S.}
}

@article{Rodero2021,
    author = {Rodero, C. and Strocchi, M. and Marciniak, M. and Longobardi, S. and Whitaker, J. and O'Neill, M. D. and Gillette, K. and Augustin, C. and Plank, G. and Vigmond, E. J. and Lamata, P. and Niederer, S. A.},
    journal = {PLOS Computational Biology},
    title = {Linking statistical shape models and simulated function in the healthy adult human heart},
    year = {2021},
    volume = {17},
    pages = {1-28}
}

@article{Verhulsdonk2024,
    journal = {Proceedings of Machine Learning Research},
    year = {2024},
    pages = {1-22},
    title = {Shape of my heart: Cardiac models through learned signed distance functions},
    author = {Verh{\"u}lsdonk, J. and Grandits, T. and Sahli Costabal, F. and Pinetz, T. and Krause, R. and Auricchio, A. and Haase, G. and Pezzuto, S. and Effland, A.}
}

@article{LiuSchiaffini2024,
    journal = {arXiv:2402.16845},
    year = {2024},
    title = {Neural Operators with Localized Integral and Differential Kernels},
    author = {Miguel Liu-Schiaffini, M. and Berner, J. and Bonev, B. and Kurth, T. and Azizzadenesheli, K. and Anandkumar, A.}
}

\end{document}